%% file: main.tex
\documentclass{article}
\usepackage{iclr2027_conference,times}

\iclrfinalcopy

\usepackage[utf8]{inputenc}
\usepackage[T1]{fontenc}
\usepackage{booktabs}
\usepackage{amsfonts}
\usepackage{amsmath}
\usepackage{amssymb}
\usepackage{microtype}
\usepackage[table]{xcolor}
\usepackage{graphicx}
\usepackage{fontawesome5}
\usepackage{xspace}
\usepackage{caption}
\usepackage{float}
\usepackage{wrapfig}
\usepackage{needspace}
\usepackage{array}
\usepackage{multirow}
\usepackage{hyperref}
\usepackage{url}
\usepackage{listings}
  \usepackage[most]{tcolorbox}
  \definecolor{promptback}{gray}{0.98}
  \tcbset{
      colback=promptback,
      colframe=black!45,
      boxrule=0.5pt,
      fonttitle=\bfseries\small,
      coltitle=white,
      left=3pt, right=3pt, top=3pt, bottom=3pt,
  }
\definecolor{citecolor}{RGB}{10, 151, 176}  
\definecolor{rewardauditbg}{RGB}{237,246,246}
\hypersetup{
    citecolor=citecolor,  
    linkcolor=citecolor,
    urlcolor=citecolor,
    colorlinks=true,
    pdftitle={When Interfaces Speak: Data-Aware Generative UI Harness for Active Interaction},
    pdfauthor={Xiaolong Li, Xiaohan Xu, Jinyang Li, Xinnuo Xu, Ge Qu, Nan Huo, Jack Williams, Reynold Cheng}
}

\newcommand{\method}{\textsc{GenUI}\xspace}
\newcommand{\bench}{\textsc{UI-Tau} Bench\xspace}

\title{When Interfaces Speak: Data-Aware\\Generative UI Harness for Active Interaction}
\author{\normalfont\parbox{0.96\textwidth}{\centering
\textbf{Xiaolong Li}$^{1,3}$ \quad
\textbf{Xiaohan Xu}$^{1,3}$ \quad
\textbf{Jinyang Li}$^{1,3}$ \quad
\textbf{Xinnuo Xu}$^{2}$ \\[0.45em]
\textbf{Ge Qu}$^{1,3}$ \quad
\textbf{Nan Huo}$^{1,3}$ \quad
\textbf{Jack Williams}$^{2}$ \quad
\textbf{Reynold Cheng}$^{1,3}$ \\[0.60em]
$^{1}$The University of Hong Kong \quad
$^{2}$Microsoft Research \\[0.20em]
$^{3}$The BIRD Team \\[0.50em]
\texttt{\{xia01ong,xxh24,jl0725\}@connect.hku.hk} \\[0.25em]
{\color{citecolor}\faGithub\ \href{https://github.com/bird-bench/GenUI-Agent}{\texttt{https://github.com/bird-bench/GenUI-Agent}}}
}}

\begin{document}

\maketitle
\lhead{Preprint}

\raggedbottom

\begin{abstract}
Most human-agent interaction today remains text-based. However, natural language is often not a perfect medium for complex tasks, as it can lead to cognitive overload, ambiguity, information chaos, and a slow input process; ephemeral generative UIs can present structured information and guide users toward task completion. In this work, we propose \textbf{\method-Harness}, a multi-agent harness that pairs a Tool Agent for information retrieval and task execution with a GUI Coder Agent that identifies ambiguities and generates front-end code for structured interfaces, helping users complete their tasks. Training such a versatile coder with reinforcement learning is challenging: verifiable rewards for interactive UI generation require costly execution, while LLM-as-a-Judge rewards are prone to reward hacking. We address the first challenge with \textbf{Dynamic UX}, a lightweight package that supports dynamic interaction and reward collection within a single sandbox, and the second with \textbf{Reward Auditor}, a meta-reward mechanism that monitors reward distributions and automatically distills diagnostic patterns into a shared rubric and scoring specification. To evaluate this task, we introduce \textbf{\bench}, a benchmark for active human-agent interaction through generated UI code, built on \textbf{10} real-world domain databases constructed from public data sources and based on Tau-Bench tool use settings, with Lite (\textbf{300 tasks}) and Full (\textbf{1,000 tasks}) splits. Experiments show that \method-Harness achieves an average Pass@3 gain of \textbf{4.48 percentage points} over smolagents on Lite. Training with \method-Harness improves a 4B backbone from a \textbf{9.33\%} baseline to \textbf{58.00\%} Pass@3, even outperforming larger frontier models such as Claude Opus 5 (\textbf{46.67\%}). We further show that GenUI-Harness remains robust not only on ambiguous queries, but also on non-ambiguous ones. In a reviewer survey comparing communication channels, generated UIs reduce the average number of dialogue rounds from \textbf{3.4} to \textbf{1.2}. These results show that data-aware generative interfaces can support effective task completion and reduce dialogue rounds in the evaluated database-backed workflows.
\end{abstract}

\section{Introduction}

Most current human-agent systems still interact through text~\citep{yao2023react,schick2023toolformer,li2023apibank,qin2024toolllm,yao2024taubench}. While natural language is flexible, it is often not a perfect medium for complex, goal-directed tasks: users face cognitive overload when translating structured needs into text, ambiguity from underspecified requirements, information chaos from scattered preferences and database records, and a slow input process when typing complex instructions~\citep{larkin1987diagram,rao2018clarification,aliannejadi2019clarifying,zamani2020mimics}. These limitations reflect a deeper mismatch between medium and task in interactive agent settings: advancing human-agent collaboration requires moving beyond conversational text to dynamic, data-aware interfaces that simultaneously visualize grounded candidate records and capture structured user intent for downstream action execution.

\begin{figure}[!t]
    \centering
    \includegraphics[width=\textwidth]{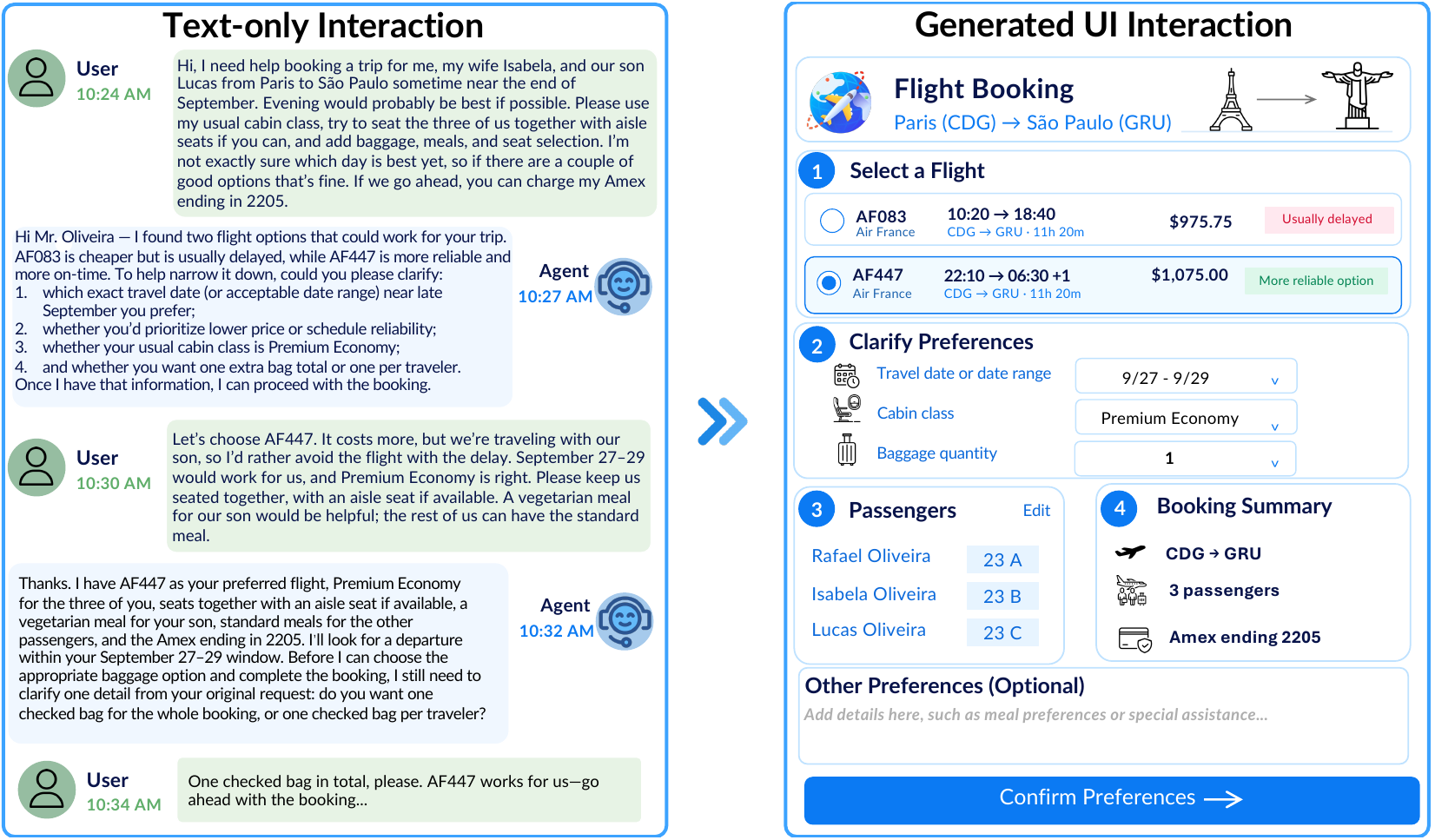}
    \captionsetup{skip=4pt}
    \caption{Text-based and generated-UI interaction for the same flight-booking request. Left: text requires successive clarification of travel preferences. Right: the generated UI presents retrieved options and structured controls for direct review and clarification.}
    \label{fig:ui-example}
    \vspace{-1.3\baselineskip}
\end{figure}

Figure~\ref{fig:ui-example} illustrates this contrast with a flight-booking request. In a conventional text interaction, flight details and clarification questions accumulate piecemeal across turns, leaving the user to piece together and evaluate competing alternatives, while the agent struggles to map fragmented user utterances into structured action arguments. This back-and-forth dialogue increases cognitive burden and the likelihood of grounding errors. In contrast, a data-aware generative interface dynamically consolidates retrieved flight candidates, pre-grounded constraints, and missing slots into a structured canvas. By anchoring candidate options directly in verified database records and restricting interactive controls to unresolved fields, the interface eliminates ambiguity at the source and collapses multi-turn textual interrogation into a single, verifiable submission ready for downstream action execution.

\Needspace{3\baselineskip}
In this paper, we study \emph{data-aware generative UI} as an alternative interaction paradigm. Building on personalized, task-specific, and dynamically synthesized interfaces~\citep{gajos2010supple,vaithilingam2019bespoke,vaithilingam2024dynavis,nandy2024bespoke,cao2025malleableui,chen2026generativeinterfaces}, an \emph{adaptive and ephemeral} interface provides a communication channel tailored to the current request and relevant database records. It presents structured information and supports direct input, following the principles of direct manipulation and focused clarification~\citep{larkin1987diagram,shneiderman1983direct,hutchins1985direct,horvitz1999mixedinitiative,xu2019clarification,aliannejadi2021clarifying,zamani2020mimics}. To realize this idea, we propose \textbf{\method-Harness}, a multi-agent harness with a Tool Agent operating in two stages and a GUI Coder Agent between them. During \emph{pre-agent exploration}, the Tool Agent gathers relevant records and candidate options through read-only tool calls. The GUI Coder Agent uses this evidence to identify ambiguities in task requirements and generates executable front-end TypeScript code that presents the retrieved options and collects the required user input. During \emph{post-agent execution}, the Tool Agent combines the submitted data with the retrieved evidence, obtains any remaining database-dependent information, and executes the requested action through tool calling.

The GUI Coder Agent must identify ambiguous task requirements, express them through appropriate interface controls, and generate code that renders correctly and submits useful data for downstream execution~\citep{wu2024uicoder,chen2023codet}. We train the coder with reinforcement learning, for which reward design poses two challenges. First, collecting execution-based rewards requires rendering generated code at scale, with isolated sessions so that one rollout does not affect another. We introduce \textbf{Dynamic UX}, a lightweight package for parallel, session-isolated execution and reward collection within a single sandbox. Second, rendering alone does not establish whether an interface exposes the relevant ambiguities or enables successful downstream action. Verifying these properties requires user interaction and task execution, while a less expensive LLM-as-a-Judge reward may assign high scores to interfaces that fail during use and is susceptible to reward hacking~\citep{zheng2023judging,tan2025judgebench,gao2023rewardoveroptimization,skalse2022rewardgaming}. We introduce \textbf{Reward Auditor}, a meta-reward mechanism that uses actual task-execution outcomes to diagnose and automatically revise the training reward. It monitors reward distributions and examines individual rollouts to identify failures that receive high rewards and successful interactions that the criteria undervalue.

To evaluate this task, we introduce \textbf{\bench}, a benchmark based on the tool-use settings of Tau-Bench~\citep{yao2024taubench} that evaluates an agent's ability to generate task-specific UIs on the fly and use them for active human-agent interaction. It grounds tasks in \textbf{10 real-world domain databases} constructed from public data sources. Its Lite (300 tasks) and Full (1{,}000 tasks) splits evaluate whether the generated interface supports the user input and downstream action needed to reach the target database state. On Lite, \method-Harness attains higher Pass@3 than \texttt{smolagents}, \texttt{mini-swe-agent}, and \texttt{Pi} on 8, 7, and 8 of the 9 shared backbones, respectively. Training a 4B backbone within our harness increases Pass@3 from 9.33\% to 58.00\%, even outperforming substantially larger frontier models such as Claude Opus 5 (46.67\%). These gains are not limited to ambiguous user requests. We also find that \method-Harness remains robust when user goals are already clearly specified, suggesting that generated interfaces are useful not only for clarification, but also for efficient task progression. In a reviewer survey comparing communication channels, generated UIs reduce average dialogue rounds from 3.4 to 1.2, showing that structured submissions can consolidate multiple required inputs into a single exchange.

Our contributions are threefold: (1) the task of \emph{data-aware generative UI} for stateful service tasks; (2) \textbf{\bench}, which evaluates active human-agent interaction through generated front-end code; and (3) \textbf{\method-Harness}, which combines tool grounding, role-specialized UI generation, Dynamic UX, and Reward Auditor. Experiments show strong database-state success for a compact model and GUI-Coder transfer across models.

\section{Task Definition}
\label{sec:problem-definition}

The data-aware generative UI task requires an agent to construct an executable interface from a user request and database evidence, collect the user input needed for the task, and complete the requested action through tool use. Each instance contains a user request, an initial database state, query and action tools, and domain rules, following the tool-use setting of Tau-Bench~\citep{yao2024taubench}. The agent uses read-only tools to retrieve relevant records from the database, generates a task-specific UI on the fly to obtain missing user information or confirm the intended action, and calls an action tool to update the database while satisfying the user's requirements and the applicable rules. Appendix~\ref{app:task-formalization} gives the formal definition.

\section{\bench}
\label{sec:bench-construction}

\bench evaluates whether an agent can generate a useful interface on the fly from the current request and database evidence, collect user input, and complete the task through tool use.

\begin{figure}[!b]
    \centering
    \includegraphics[width=\textwidth]{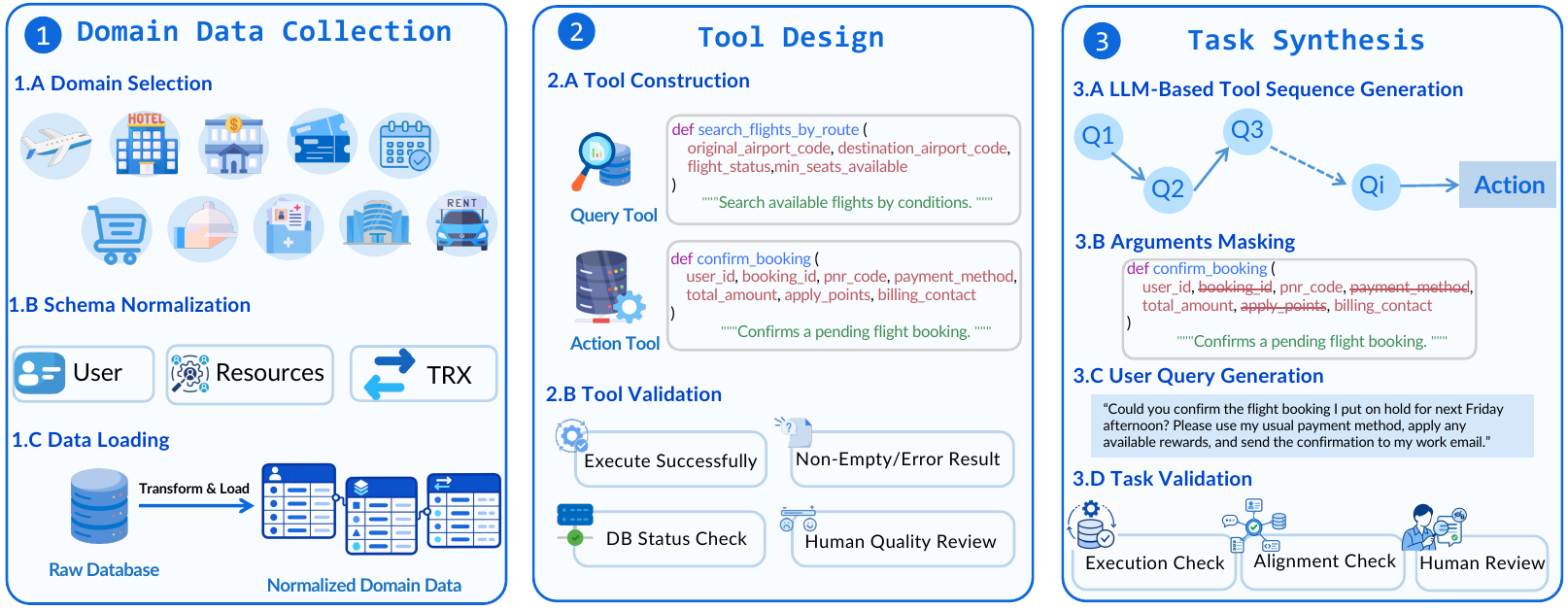}
    \captionsetup{skip=4pt}
    \caption{\bench construction pipeline. We collect and normalize the domain databases, construct and validate the query and action tools, and synthesize and validate the benchmark tasks.}
    \label{fig:ui_bench}
\end{figure}

\paragraph{Domain Data Collection.} The benchmark uses 10 real-world domain databases constructed from public data sources, covering flight booking, hotel booking, e-commerce, and retail~\citep{openflightsdata,antonio2019hotel,olist2018brazilian,chen2012onlineretailii}. These sources provide domain records in heterogeneous schemas, rather than task environments with a common query interface and defined state transitions. Four doctoral researchers collect and normalize these sources by standardizing field types and identifiers, checking cross-table references, and organizing each domain into a user-facing entity table, resource tables, and a transaction table. Query tools can then retrieve related records, and action tools can apply updates whose effects are checked against a target database state under explicit domain rules. Appendix~\ref{app:construction} documents the data sources and construction details.

\paragraph{Tool Design.} Working from the normalized databases, the same annotators first specify the domain rules that govern action requirements, eligibility conditions, and valid state transitions, then define the corresponding query and action tools~\citep{li2023apibank,qin2024toolllm}. Query tools are atomic, read-only retrieval operations that compose into search chains. Action tools apply rule-compliant updates whose arguments are grounded in query observations or user clarification. Validation applies a separate condition to each tool type. A query tool must execute without error, return a non-empty result, and return records consistent with the arguments it was given. An action tool is invoked with arguments taken from query observations or user clarification and must both satisfy the domain rules and produce the intended database state. A tool that violates any condition is revised against the case that failed and revalidated at most twice. Every surviving tool is then cross-checked by a second annotator. We retain a tool when both annotators agree that it is correct, and discard tools that still fail or whose behavior remains ambiguous. Appendix~\ref{app:construction-pipeline} details tool construction and validation.

\begin{wraptable}[12]{r}{0.42\textwidth}
    \vspace{-1.0\baselineskip}
    \centering
    \captionsetup{skip=4pt}
    \caption{Data statistics}
    \label{tab:bench-summary}
    \resizebox{\linewidth}{!}{%
    \begin{tabular}{@{}lcc@{}}
        \toprule
        \textbf{\textsc{Statistic}} & \textbf{\textsc{Lite}} & \textbf{\textsc{Full}} \\
        \midrule
        \textbf{Total instances}       & \textbf{300} & \textbf{1,000} \\
        Moderate-ambiguity instances   & 161          & 523 \\
        High-ambiguity instances       & 139          & 477 \\
        Unique query tools             & 303          & 304 \\
        Unique action tools            & 162          & 331 \\
        Avg. tokens / user task        & 90.3         & 90.2 \\
        Avg. masked args. / action     & 4.9          & 4.8 \\
        \midrule
        Inter-annotator agreement (\%) & 94.1         & 94.3 \\
        \bottomrule
    \end{tabular}}
\end{wraptable}
\paragraph{Task Synthesis.} We design a task synthesis agent using \texttt{Claude Opus 5}. For each candidate, one of these annotators samples a random user and specifies a task goal, such as changing an existing flight booking. Given the selected user's records, the domain database, and the available tools, the agent proposes and executes a query chain for that goal, with later calls using earlier results, and constructs a reference action whose arguments come from retrieved records. We then mask all internal identifiers except the user identifier, together with 2--3 additional arguments for moderate-ambiguity tasks or 4--5 for high-ambiguity tasks. Conditioned on the executed chain, retrieved records, and domain rules, the agent writes a natural-language request that leaves these arguments unspecified. The hidden reference action preserves the complete goal, while the visible request requires the evaluated agent to recover missing information through database exploration, user interaction, or both. Each candidate passes three checks: execution verifies the reference action's database update; alignment checks establish consistency among the request, user profile, records, rules, and reference action; and two independent annotators assess plausibility, fluency, and masked-value leakage. Disagreements are resolved by a third annotator, and unresolved candidates are discarded. Appendix~\ref{app:construction-pipeline} gives the full protocol.

\paragraph{Data Statistics and Evaluation Metrics.} Table~\ref{tab:bench-summary} shows the \bench statistics. Appendix~\ref{app:bench-stats} reports per-domain statistics and comparisons with existing benchmarks. We conduct three evaluation sessions per task and report state exact-match (EM) metrics over $N$ tasks:
\[
\mathrm{Pass@3} = \frac{100}{N}\sum_{i=1}^{N}\max_{j\in\{1,2,3\}}z_{i,j},
\qquad
\mathrm{Avg@3} = \frac{100}{3N}\sum_{i=1}^{N}\sum_{j=1}^{3}z_{i,j}.
\]
Here, $z_{i,j}=1$ if session $j$ executes successfully and produces exactly the reference database state, and $0$ otherwise. Pass@3 measures success in at least one session; Avg@3 averages success across all three.

\section{\textbf{\method-Harness}}
\label{sec:genui-harness}

\begin{figure}[htbp]
    \centering
    \includegraphics[width=0.8\textwidth]{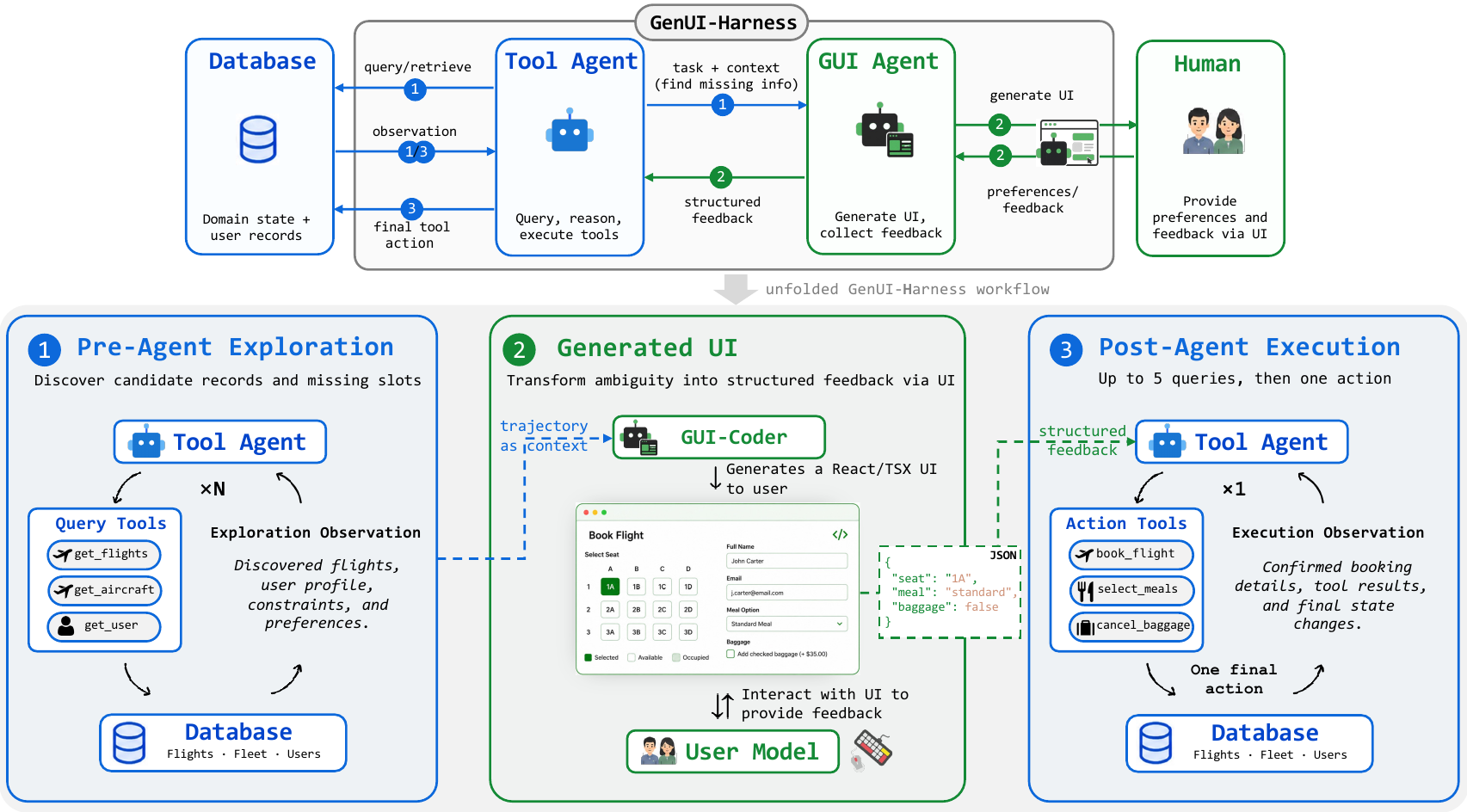}
    \captionsetup{skip=4pt}
    \caption{\method-Harness links three stages: the Tool Agent retrieves task context, the GUI Coder Agent generates an interface to collect user input, and the Tool Agent uses that input to execute the final database action.}
    \label{fig:genui_agent}
\end{figure}

Solving each task in \bench requires the agent system to understand the query, explore the database, and interact with the user. \method-Harness therefore assigns these requirements to separate roles. A Tool Agent retrieves records and executes the final action, and a GUI Coder Agent turns the retrieved evidence and the unresolved arguments into an interface the user operates directly, so that choices are made among grounded options without exposing database schemas or internal identifiers. Separating the roles also makes the interface component independently trainable, which allows the GUI Coder Agent to be optimized against execution outcomes while the Tool Agent is held fixed, and allows the trained coder to be paired with a different tool-reasoning model. Figure~\ref{fig:genui_agent} summarizes the workflow, and Appendix~\ref{app:agent-workflow} gives the detailed protocols.

\paragraph{Pre-Agent Exploration.} Given the query and candidate action schemas, the Tool Agent identifies the intended operation and uses read-only tools in a ReAct loop~\citep{yao2023react,schick2023toolformer,li2023apibank} (up to 5 turns) to retrieve records and the user profile. This grounds known arguments and collects candidate options for unresolved user choices. The Tool Agent passes the query and the exploration history, which records its tool calls and their observations, to the GUI Coder Agent.

\paragraph{Generated UI.} The GUI Coder Agent reads the query and the exploration history and determines which arguments the retrieved records already fix, which candidate values are available for the remaining ones, and which arguments only the user can decide. For each argument left to the user it selects an interface control suited to that decision, such as a list for choosing among retrieved candidates or an input field for a value the user supplies~\citep{horvitz1999mixedinitiative,vaithilingam2024dynavis,ma2025ambigchat}. It emits a single Material UI component in React and TypeScript that presents the candidates with readable labels and the attributes a decision depends on. Arguments already resolved are shown as context or prefilled, so the user attends only to what is unresolved. The user simulator operates the rendered interface according to its assigned intent and preferences, and its submission is returned to the Tool Agent as a JSON object whose keys name action arguments and whose values record the user inputs.

\paragraph{Post-Agent Execution.} The Tool Agent resumes with the query, exploration history, and submitted user inputs. It can use up to 5 further query turns to validate selections or recover remaining database-dependent arguments. It combines these observations with the user's input to complete the action arguments, then executes the action tool to update the database.

\section{Training}
\label{sec:training}
We construct training environments from 40 databases synthesized with \texttt{Claude Sonnet 4.6}~\citep{anthropic2026sonnet46}, all separate from \bench. On all 40 databases, \texttt{DeepSeek V4 Flash}~\citep{xu2026deepseek} generates end-to-end trajectories within \method-Harness. We retain successful trajectories whose final database state matches the target and whose interfaces render with valid ARIA snapshots. Trajectories from 25 databases are used to fine-tune the Tool Agent and GUI Coder Agent separately with supervised fine-tuning (SFT). The remaining 15 databases define the RL environments; retained trajectories from these databases supply reference UI code and ARIA snapshots for reward evaluation. Appendix~\ref{app:training} details construction, filtering, and split verification.

Both agents start from \texttt{Qwen3.5-4B} and are fine-tuned separately on their respective training examples, which fixes the trajectory structure and the output format each agent must produce. We call the resulting pair of checkpoints \method-4B (SFT). Holding the Tool Agent at its SFT checkpoint, we then train only the GUI Coder Agent on the 15 RL databases with Group Relative Policy Optimization (GRPO)~\citep{shao2024deepseekmath}, using the audited reward to optimize it toward interfaces that lead to successful task completion~\citep{le2022coderl,wu2024uicoder}.

\subsection{Dynamic UX}
Static code checks cannot establish whether a generated UI renders successfully or what controls it exposes. The component must be transpiled, loaded, and executed in a browser to obtain the interface evidence used for reward evaluation. Repeating this process for every rollout adds rendering time before policy updates. A straightforward isolation strategy launches a fresh browser or sandbox per rollout, duplicating browser processes and, with separate sandboxes, servers and dependencies as concurrency grows. Dynamic UX instead shares one rendering runtime and browser process within a single sandbox while assigning each rollout an isolated browser context, so execution results and rewards cannot be affected by other samples.

Dynamic UX initializes the web server, socket router, browser, and component library once, then assigns each rollout a unique session and isolated browser context. It returns execution status and an initial ARIA snapshot for reward collection, while session-based routing prevents state from crossing samples. To evaluate rendering efficiency and scalability, we benchmark Dynamic UX on 1{,}100 pre-verified components against fresh-process rendering across five concurrency levels. Results show that context-level isolation is consistently more efficient: on average, Dynamic UX increases throughput by 18.2\% while reducing median latency by 15.3\%, peak resident set size (RSS) by 35.2\%, and Chromium process count by 59.4\%. A 1{,}000-session marker test completes without timeout, failure, or content leakage. Appendix~\ref{app:dynamic-ux} reports the full setup and results.

\subsection{Reward Auditor}
\label{sec:reward-auditor}
Successful rendering does not establish that an interface enables task completion. Verifying the latter requires user interaction, downstream action execution, and a check of the resulting database state, making full evaluation costly for every training rollout. A language-model judge provides less expensive feedback, but may reward plausible-looking interfaces that fail during use or undervalue useful alternatives~\citep{zheng2023judging,kim2024prometheus2,lambert2025rewardbench,tan2025judgebench,gao2023rewardoveroptimization}. Reward Auditor addresses this gap with a meta-reward mechanism that uses task-execution outcomes to diagnose and iteratively revise the reward that trains the GUI Coder Agent.

We construct the initial reward from task requirements and successful reference trajectories. The judge compares the generated code and rendered ARIA snapshot with the user query, reference code, and reference ARIA. Its rubric operationalizes four requirements: functional equivalence, feature completeness, implementation correctness, and task alignment, with scores $q_i^{\mathrm{fe}},q_i^{\mathrm{fc}},q_i^{\mathrm{ic}},q_i^{\mathrm{ta}}\in[0,5]$. Let $g_i$ indicate that the rollout passes the required code, rendering, reference-evidence, and judge-response checks. The initial execution-gated reward is
\[
j_i=\frac{0.35q_i^{\mathrm{fe}}+0.35q_i^{\mathrm{fc}}+0.15q_i^{\mathrm{ic}}+0.15q_i^{\mathrm{ta}}}{5},
\qquad r_i^{(0)}=g_i j_i.
\]
This initial rubric provides an interpretable starting point without requiring exact code agreement with the reference. Its specification appears in Appendix~\ref{app:initial-reward}.

This initial reward assigns the same zero score to failures in code extraction, structural validation, and rendering. If every output in a prompt group fails these checks, the rewards provide no within-group contrast for GRPO. Conversely, a rendered interface can receive a high judge score even when its controls cannot submit the user's intended values. Auditing therefore needs to address both the judge's criteria and the mapping from execution evidence to scalar rewards.

\begin{wrapfigure}[20]{r}{0.42\textwidth}
    \vspace{-0.65\baselineskip}
    \centering
    \includegraphics[width=\linewidth]{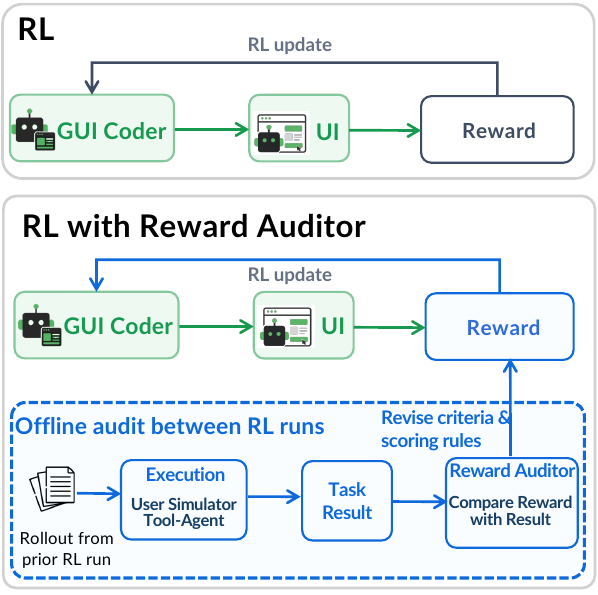}
    \captionsetup{font=normalsize,skip=3pt}
    \caption{Reward Auditor adds an offline reward-revision loop to RL. Task outcomes on selected internal rollouts guide revisions to judging criteria and scoring rules.}
    \label{fig:reward-auditor}
\end{wrapfigure}

Reward Auditor uses \texttt{Claude Opus 5}~\citep{anthropic2026opus5} to diagnose and revise rewards from task-execution evidence on internal rollouts from the 15 RL databases (Figure~\ref{fig:reward-auditor}). The user simulator operates the interface, and the fixed Tool Agent consumes the submission and executes a terminal action.

\Needspace{20\baselineskip}
The objective label $z_i$ is positive only when execution succeeds and the resulting database state exactly matches the target. For each audit round, Reward Auditor compares this label with the reward prediction $\hat{z}_i=\mathbb{I}[r_i\geq0.5]$ and examines the resulting cases, reward distributions, judge rationales, and complete trajectories to diagnose failure causes and recurring reward--outcome mismatches.

The auditor then modifies the judging criteria and scoring rules and reviews the proposed changes. It replays candidate rewards on stored rollouts and reviews the resulting scores before accepting a revision. Replay holds generated code fixed, making score changes attributable to the reward revision rather than policy behavior. Task-specific diagnoses are consolidated into one shared rubric and scoring specification applied across tasks. The revisions address interaction and submitted-payload requirements, distinguish failure stages, and introduce targeted icon and accessibility checks. We perform two rounds of meta-reward-guided revision, producing the final training reward $r_i^{(2)}$ used for the reported \method-4B RL session. In our experiments, Direct RL and both audited variants train independent policies from the same SFT initialization with the Tool Agent fixed.

The final reward separates code-extraction, structural, rendering, and empty-interface failures before applying quality and accessibility scoring. Appendix~\ref{app:reward-design} gives its complete specification and revision protocol.

\section{Experiments}
\subsection{Setup}
We benchmark four agent harnesses on \bench: \method-Harness, \texttt{smolagents}~\citep{smolagents2025}, \texttt{mini-swe-agent}~\citep{minisweagent2025}, and \texttt{Pi}~\citep{piagent2026}. All four are evaluated on the same tasks, databases, query and action tools, user-simulator model, and final-state evaluator, with the same backbone in each row. \method-Harness communicates through generated interfaces, whereas the three baselines retain their original text-based interaction. Model and inference details appear in Appendix~\ref{app:inference-config}, and harness adaptations in Appendix~\ref{app:harness-adaptation}. We report Pass@3 and Avg@3 as defined in Section~\ref{sec:bench-construction}.

\subsection{Main Results}

\paragraph{Overall Performance.}
Table~\ref{tab:main} presents the end-to-end Lite comparison. The three baselines use their native text interaction, providing the text-only control for our generated-UI system while sharing the task environment, tool APIs, user-simulator model, and evaluator; no GUI Coder Agent or browser renderer is inserted into their paths. \method-Harness leads Pass@3 on seven of the nine backbones across open-weight and closed-source models. Macro-averaged across these backbones, it improves Pass@3 over \texttt{smolagents} by 4.48 percentage points (task-paired bootstrap 95\% CI: [2.37, 6.59]); Table~\ref{tab:lite-bootstrap} reports uncertainty against all three baselines. Because the comparison preserves each harness's native agent loop and communication mode, it measures complete system performance rather than isolating the interface channel alone; the reviewer study in Section~\ref{sec:human-eval} compares the two communication channels directly. Full-split results are presented in Appendix~\ref{app:full-results}.

\begin{figure}[t]
    \centering
    \begin{minipage}[t]{0.50\textwidth}
        \centering
        \includegraphics[height=0.145\textheight,keepaspectratio]{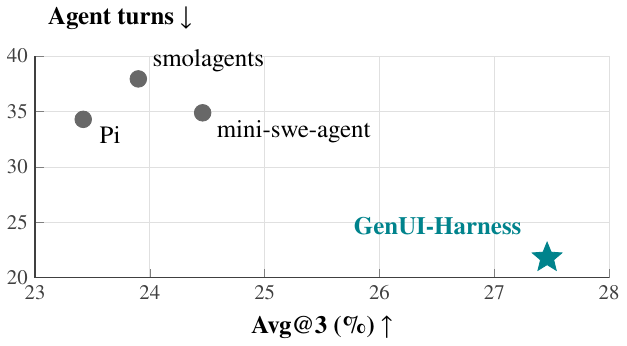}
        \captionsetup{font=small,skip=3pt}
        \caption{Avg@3 versus agent turns on Lite.}
        \label{fig:harness-efficiency}
    \end{minipage}%
    \begin{minipage}[t]{0.50\textwidth}
        \centering
        \raisebox{-2.5pt}{\includegraphics[height=0.145\textheight,keepaspectratio]{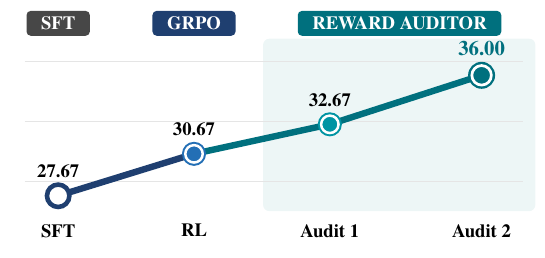}}
        \captionsetup{font=small,skip=3pt}
        \caption{Avg@3 (\%) across GUI Coder training stages.}
        \label{fig:gui-coder-rl-progression}
    \end{minipage}
\end{figure}

\paragraph{Interaction Efficiency.}
Figure~\ref{fig:harness-efficiency} places \method-Harness alone on the favorable lower-right frontier: it achieves higher Avg@3 with 36--43\% fewer agent turns among successful episodes. This suggests that the harness avoids redundant reasoning and recovery: the Tool Agent focuses on retrieval and validation, while the generated UI collects multiple grounded choices in one structured interaction.

\begin{table}[t]
    \centering
    \caption{State EM Pass@3 and Avg@3 (\%) on the 300-task Lite split across four harnesses (three sessions; temperature 0.7). \method-Harness uses generated UI; the other harnesses retain native text interaction. Bold marks each row's best result; ties are bolded jointly.}
    \label{tab:main}
    \small
    \setlength{\tabcolsep}{5pt}
    \renewcommand{\arraystretch}{1.04}
    \resizebox{\textwidth}{!}{%
    \begin{tabular}{@{}lcccccccc@{}}
        \toprule
        & \multicolumn{2}{>{\columncolor{rewardauditbg}}c}{\textbf{\method-Harness}} & \multicolumn{2}{c}{\textbf{smolagents}} & \multicolumn{2}{c}{\textbf{mini-swe-agent}} & \multicolumn{2}{c}{\textbf{Pi}} \\
        \cmidrule(lr){2-3} \cmidrule(lr){4-5} \cmidrule(lr){6-7} \cmidrule(lr){8-9}
        \textbf{Model} & \cellcolor{rewardauditbg}\textbf{Pass@3} & \cellcolor{rewardauditbg}\textbf{Avg@3} & \textbf{Pass@3} & \textbf{Avg@3} & \textbf{Pass@3} & \textbf{Avg@3} & \textbf{Pass@3} & \textbf{Avg@3} \\
        \midrule
        \rowcolor{black!4}
        \multicolumn{9}{@{}l}{\hspace{0.45em}\textsc{Open-weight models}} \\
        \addlinespace[1pt]
        \texttt{Qwen3.5-4B}        & \cellcolor{rewardauditbg} 9.33 & \cellcolor{rewardauditbg} 3.22 & 10.67 &  3.90 & 11.67 &  4.11 & \textbf{13.67} & \textbf{5.00} \\
        \texttt{Qwen3.6-27B}       & \cellcolor{rewardauditbg}\textbf{54.33} & \cellcolor{rewardauditbg}\textbf{32.44} & 51.67 & 30.20 & 43.33 & 27.67 & 43.33 & 28.00 \\
        \texttt{Qwen3.6-35B-A3B}   & \cellcolor{rewardauditbg}\textbf{47.33} & \cellcolor{rewardauditbg}\textbf{26.89} & 43.67 & 25.90 & 36.33 & 20.11 & 38.00 & 21.78 \\
        \texttt{gpt-oss-120b}      & \cellcolor{rewardauditbg}47.67 & \cellcolor{rewardauditbg}\textbf{28.22} & 30.00 & 14.10 & \textbf{50.00} & \textbf{28.22} & 40.33 & 23.11 \\
        \texttt{DeepSeek V4 Flash} & \cellcolor{rewardauditbg}\textbf{49.67} & \cellcolor{rewardauditbg}\textbf{30.22} & 49.33 & 29.70 & 42.00 & 27.22 & 40.67 & 25.78 \\
        \texttt{Qwen3.5-397B-A17B} & \cellcolor{rewardauditbg}\textbf{56.00} & \cellcolor{rewardauditbg}\textbf{33.67} & 48.33 & 32.90 & 43.33 & 28.22 & 42.33 & 27.89 \\
        \addlinespace[4pt]
        \rowcolor{black!4}
        \multicolumn{9}{@{}l}{\hspace{0.45em}\textsc{Closed-source models}} \\
        \addlinespace[1pt]
        \texttt{GPT-5.4}           & \cellcolor{rewardauditbg}\textbf{40.00} & \cellcolor{rewardauditbg}\textbf{25.22} & 35.67 & 18.90 & 35.67 & 23.00 & 35.67 & 22.11 \\
        \texttt{Claude Sonnet 4.6} & \cellcolor{rewardauditbg}\textbf{51.33} & \cellcolor{rewardauditbg}\textbf{35.67} & 48.00 & 31.30 & 47.67 & 33.56 & 40.67 & 28.44 \\
        \texttt{Claude Opus 5}     & \cellcolor{rewardauditbg}\textbf{46.67} & \cellcolor{rewardauditbg}\textbf{31.56} & 44.67 & 28.20 & 40.00 & 28.00 & 43.33 & 28.67 \\
        \bottomrule
    \end{tabular}%
    }
\end{table}

\paragraph{Role-Specialized Training and Reward Auditing.}
We further train a compact \texttt{Qwen3.5-4B} within \method-Harness through role-specific supervision and audited rewards. SFT teaches the division of labor: the Tool Agent retrieves evidence and executes actions, while the GUI Coder Agent turns unresolved requirements into executable interfaces, increasing Avg@3 from 3.22\% to 27.67\%. We then isolate the contribution of RL by fixing the SFT Tool Agent and GUI-Coder initialization (Figure~\ref{fig:gui-coder-rl-progression}).
Dynamic UX executes generated code and collects rollout rewards for GUI-Coder GRPO at scale. The initial judge-based reward can favor plausible interfaces with broken controls or payload semantics and assign indistinguishable scores to different failure stages. Reward Auditor compares these rewards with exact database-state outcomes, diagnoses distribution-level patterns and individual cases, and automatically revises the shared rubric and scoring rules through replay. Replay holds sampled interfaces fixed, separating changes in reward behavior from changes in policy outputs. The final reward distinguishes code-extraction, structural, rendering, and empty-interface failures before applying quality and accessibility scoring. With the final audited reward, \method-4B reaches 58.00\% Pass@3 and 36.00\% Avg@3, the best result in Figure~\ref{fig:gui-coder-rl-progression} and above the strongest Table~\ref{tab:main} baseline on both metrics.
\subsection{Analysis}

\paragraph{Error Analysis.}
Across four models on the 300 Lite tasks, failures are dominated by unsuccessful UI interaction (52.0\%) and mismatched action arguments (44.5\%). Trace inspection identifies UI defects that prevent users from expressing required selections or submitting correct values; state mismatches require examining both submissions and resulting database changes. For example, a single-select control cannot express a required multi-selection, and an immutable cabin-class field submits the wrong value despite a fully specified user goal. These findings highlight interface operability and argument fidelity as key bottlenecks. Appendix~\ref{app:error-analysis} provides detailed counts, diagnostic procedures, and case studies.

\Needspace{7\baselineskip}
\paragraph{Communication-Channel Analysis.}
\label{sec:human-eval}
We compare channels on 100 tasks whose interfaces execute and support valid user-simulator interaction in all three sessions ($\mathrm{pass}^{3}$). For each task, three reviewers see the UI and three see a text rendering of the same task state; assignments are randomized, and each reviewer sees only one channel. Generated UIs reduce average dialogue rounds from \textbf{3.4} to \textbf{1.2}. UI latency is judged acceptable on \textbf{85\%} of tasks by a majority of UI reviewers. Appendix~\ref{app:human-study} details task selection, channel construction, latency measurement, and the evaluation protocol.

\par\medskip
\Needspace{13\baselineskip}
\noindent
\begin{minipage}[t]{0.48\textwidth}
    \vspace{0pt}
    \paragraph{Code-Stage Intervention Across Models.}
    To study the generalizability of GUI-Coder, we pair it with five other reasoning models within \method-Harness. For each model, we hold its original pre-agent output fixed, replace only code generation with GUI-Coder, collect a new UI, and rerun the same model as the post-agent. Table~\ref{tab:frontier-gui-coder} shows higher Lite Avg@3 for all five models. Because the upstream exploration trace and post-agent backbone are fixed, the intervention changes only interface construction. The consistent improvements therefore indicate that GUI-Coder adapts to distinct reasoning models rather than relying on a particular upstream agent.
\end{minipage}\hfill%
\begin{minipage}[t]{0.50\textwidth}
    \vspace{0pt}
    \centering
    \captionsetup{font=small,skip=3pt}
    \captionof{table}{Avg@3 (\%) for code-stage intervention on Lite, reusing the same pre-agent output. Green arrows show absolute gains.}
    \label{tab:frontier-gui-coder}
    \begingroup
    \small
    \setlength{\tabcolsep}{2.5pt}
    \renewcommand{\arraystretch}{1.0}
    \begin{tabular}{@{}lcc@{}}
        \toprule
        & \multicolumn{2}{c}{\textbf{Avg@3 (\%) $\uparrow$}} \\
        \cmidrule(lr){2-3}
        \textbf{Model} & \textbf{Original} & \textbf{+ GUI-Coder} \\
        \midrule
        \texttt{Qwen3.6-35B-A3B} & 26.89 & \textbf{29.33}~\textcolor{green!50!black}{\scriptsize($\uparrow$2.44)} \\
        \texttt{DeepSeek V4 Flash} & 30.22 & \textbf{38.67}~\textcolor{green!50!black}{\scriptsize($\uparrow$8.45)} \\
        \texttt{GPT-5.4} & 25.22 & \textbf{30.00}~\textcolor{green!50!black}{\scriptsize($\uparrow$4.78)} \\
        \texttt{Claude Sonnet 4.6} & 35.67 & \textbf{42.33}~\textcolor{green!50!black}{\scriptsize($\uparrow$6.66)} \\
        \texttt{Claude Opus 5} & 31.56 & \textbf{40.33}~\textcolor{green!50!black}{\scriptsize($\uparrow$8.77)} \\
        \bottomrule
    \end{tabular}
    \endgroup
\end{minipage}
\par

\par\medskip
\Needspace{14\baselineskip}
\noindent
\begin{minipage}[t]{0.48\textwidth}
    \vspace{0pt}
    \paragraph{Evaluation on Unambiguous Tasks.}
    In some scenarios, the user query states the intent clearly. To analyze \method-Harness performance in this setting, we create non-ambiguous Lite variants by revealing the user-facing arguments masked in the original requests while keeping the tools, databases, user simulator, and evaluation protocol fixed. Figure~\ref{fig:ambiguity-robustness} shows that \method-4B remains strongest in both settings. We find that the UI provides guided task progression by organizing retrieved information for review and supporting confirmation of the intended action.
\end{minipage}\hfill%
\begin{minipage}[t]{0.50\textwidth}
    \vspace{0pt}
    \centering
    \includegraphics[width=\linewidth]{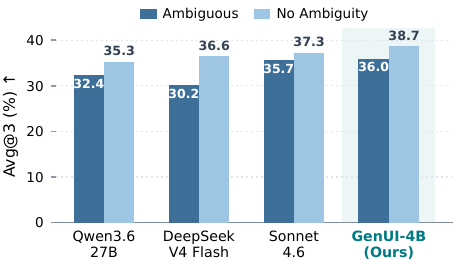}
    \captionsetup{font=normalsize,skip=4pt}
    \captionof{figure}{Avg@3 (\%) on Lite, with and without ambiguity.}
    \label{fig:ambiguity-robustness}
\end{minipage}
\par

\section{Related Work}
\label{sec:related-work}

Tool-using agents and state-based benchmarks study planning, API execution, and task completion, but generally retain language-based interaction~\citep{yao2023react,yao2024taubench}. Clarification and mixed-initiative systems recover missing intent through questions, structured options, or generated interfaces, but typically separate interface generation from database-backed action execution~\citep{horvitz1999mixedinitiative,rao2018clarification,aliannejadi2019clarifying,vaithilingam2024dynavis,chen2026generativeinterfaces}. Execution-based UI-code methods improve functional correctness through rendering or tests while treating code generation as the endpoint rather than part of an agent loop~\citep{wu2024uicoder,chen2023codet}. We instead connect retrieved records, executable interaction, and terminal state evaluation within one task. Appendix~\ref{app:related-work} provides a detailed related-work discussion.

\section{Conclusion}

\method-Harness separates database grounding and action execution from task-conditioned interface generation. Dynamic UX scales GUI rollouts, while Reward Auditor revises rewards using task outcomes. On \bench Lite, \method-4B achieves 58.00\% Pass@3 and 36.00\% Avg@3, outperforming all evaluated baselines; GUI-Coder also improves five models. Among successful episodes, \method-Harness uses 36--43\% fewer agent turns, and generated UIs reduce reviewer dialogue from 3.4 to 1.2 rounds. Together, these results support executable, data-grounded interfaces as an effective interaction layer for database-backed agents.

\subsection*{AI use statement}
Generative models are used as experimental components for synthetic benchmark construction, trajectory generation, user simulation, reward judging, and automated reward revision, as described in the paper. Generative AI tools also provided feedback on experimental protocols and result interpretation and assisted with preparing and checking the code and data preview. Separately, they were used to polish prose and formatting. The authors reviewed the AI-assisted work and take responsibility for the final text, claims, results, and artifacts.

\subsection*{Reproducibility statement}
Appendix~\ref{app:construction} documents the benchmark sources, normalization, tool validation, and task synthesis. Appendix~\ref{app:training} provides training and inference settings, the Dynamic UX benchmark, and reward specifications. Appendices~\ref{app:task-and-agent}, \ref{app:human-study}, and~\ref{app:prompts} describe the agent workflow, reviewer evaluation protocol, and prompts. The accompanying code repository contains inference and Dynamic UX components with preview data.

\bibliography{references}
\bibliographystyle{iclr2027_conference}

\clearpage
\appendix

\section{Task Formulation and Agent Workflow}
\label{app:task-and-agent}

\subsection{Formal Task Definition and Evaluation}
\label{app:task-formalization}

A task consists of a user query, an initial domain database, query and action tools with domain rules, and a specific user goal. The query may express only part of that goal. The agent observes the query, tool schemas, retrieved records, and subsequent user input. Read-only query calls gather evidence; the generated interface obtains user input; one terminal action updates the database.

Each task specifies an initial database state $s_{\mathrm{db}}^0$ and a reference action $a^\star=(\texttt{name}^\star,\mathbf{v}^\star)$, where $\mathbf{v}^\star=\{(k_j,v_j)\}_{j=1}^{n}$ contains the grounded value of every required argument. Executing $a^\star$ from $s_{\mathrm{db}}^0$ produces the target database state $s_{\mathrm{db}}^\star$ used for evaluation. For task $i$ and session $j\in\{1,2,3\}$, the evaluator resets an isolated database copy to $s_{\mathrm{db},i}^0$, executes the agent's terminal action, and observes the resulting state $\hat{s}_{\mathrm{db},i,j}$. Let $e_{i,j}=1$ indicate successful execution. The state-EM outcome is
\[
z_{i,j}=\mathbb{I}\!\left[e_{i,j}=1\ \land\ \hat{s}_{\mathrm{db},i,j}=s_{\mathrm{db},i}^\star\right].
\]
Over $N$ tasks, Pass@3 is $\frac{100}{N}\sum_i\max_j z_{i,j}$ and Avg@3 is $\frac{100}{3N}\sum_i\sum_j z_{i,j}$.
The entire resulting database state must match the target. This credits any action realization that induces the verified state transition, while failed execution and every missing, extra, or incorrect state change count as failures.

\subsection{Detailed Agent Workflow}
\label{app:agent-workflow}
The Tool Agent performs up to five read-only query turns before UI generation. The GUI Coder Agent receives the initial query and exploration history, first reflects on the remaining information needs, and then produces a single React/TypeScript component. The interface displays retrieved options and collects user input through \texttt{submitData}. The Tool Agent resumes with the query, exploration history, and submitted JSON, uses up to five further query turns to validate or resolve selections, and emits one terminal action.

The user simulator receives the generated TSX, the initial ARIA snapshot, and the complete user goal encoded as the reference action. It generates Playwright code to operate the interface according to that goal. The simulator is fixed across all harnesses and nine backbones and does not determine success. In generated-UI sessions, it cannot invoke task tools or directly modify the database; it must act through the rendered interface, after which the fixed Tool Agent consumes its submission and the evaluator independently checks the terminal database state. The reference action is private to the simulator and is not an additional input to the Tool Agent or GUI Coder Agent. The same simulator also operates interfaces produced by unrelated model families, while the reviewer study in Appendix~\ref{app:human-study} confirms that people can operate the selected interfaces under the same task information. We therefore use it as a controlled, capable-user simulator without claiming that it reproduces the full distribution of human interaction behavior. Appendix~\ref{app:prompts} gives the role templates, and Appendix~\ref{app:inference-config} specifies serving and retry settings.

\section{Related Work}
\label{app:related-work}

\paragraph{Tool-Using and Interactive Agents.}
Tool-using agents combine language reasoning with API calls to retrieve information and execute actions~\citep{yao2023react,schick2023toolformer,li2023apibank,qin2024toolllm}. Their evaluation covers general assistance and stateful environments~\citep{liu2024agentbench,mialon2023gaia,trivedi2024appworld}, as well as interaction with existing web and desktop interfaces~\citep{yao2022webshop,deng2023mindweb,zhou2023webarena,koh2024visualwebarena,drouin2024workarena,xie2024osworld}. Conversational benchmarks further study how agents respond to evolving user requests during navigation and tool use~\citep{lu2024weblinx,deng2024multiturnweb,lu2025toolsandbox}. Tau-Bench is particularly close to our setting: it combines simulated users, domain-specific tools, and database-state evaluation, with clarification conducted through textual dialogue~\citep{yao2024taubench}. We retain database-backed task execution but change how user choices are elicited: the agent generates an executable interface from retrieved records and the action schema, allowing users to inspect options and supply the choices needed for the final action.

\paragraph{Clarification and Generative Interfaces.}
Mixed-initiative systems coordinate automated inference with user control, while clarification research studies how agents identify and resolve underspecified requests~\citep{horvitz1999mixedinitiative,rao2018clarification,aliannejadi2019clarifying,xu2019clarification,min2020ambigqa,zhang2024clamber,zhang2025clarify}. MIMICS operationalizes clarification through question-and-option panes, and AmbigChat uses interactive widgets for hierarchical disambiguation~\citep{zamani2020mimics,ma2025ambigchat}. In HCI, direct manipulation and automatically generated interfaces have long been used to expose task structure and reduce the gap between intent and action~\citep{shneiderman1983direct,hutchins1985direct,gajos2010supple}. More recent systems synthesize task-specific or malleable interfaces from demonstrations, natural-language requests, or evolving data models~\citep{vaithilingam2019bespoke,vaithilingam2024dynavis,nandy2024bespoke,cao2025malleableui,chen2026generativeinterfaces}. Contemporaneous EvoGenUI-Bench, accepted to EMNLP 2026, evaluates whether one executable interface preserves requirements and tool grounding across five successive revision turns~\citep{peng2026evogenuibench}; our benchmark instead evaluates task-specific interfaces that elicit unresolved intent and complete a terminal database action. Our setting is distinguished by database-grounded service interaction: retrieved records and an executable action schema determine what the interface must clarify, and correctness is verified by executing the terminal action and exactly checking the resulting database state.

\paragraph{UI Code Generation and Outcome-Based Feedback.}
Prior work includes screenshot-based GUI automation and datasets~\citep{yeh2009sikuli,deka2017rico}, screenshot-to-code reconstruction~\citep{beltramelli2018pix2code,laurencon2024websight,si2025design2code}, and UI-code fine-tuning with compiler and multimodal feedback~\citep{wu2024uicoder}. CodeRL and CodeT demonstrate the value of execution and generated tests when reference similarity is insufficient for functional correctness~\citep{le2022coderl,chen2023codet}. Model-generated feedback has also been used for scalable alignment and self-improvement~\citep{bai2022constitutional,lee2024rlaif,yuan2024selfrewarding}, but judge and reward-model reliability remains an active concern~\citep{zheng2023judging,kim2024prometheus2,lambert2025rewardbench,tan2025judgebench}. In particular, optimizing imperfect proxies can induce overoptimization or reward gaming~\citep{gao2023rewardoveroptimization,skalse2022rewardgaming}. We combine online execution-gated feedback with a meta-reward process that uses observed task outcomes to diagnose and iteratively revise the judging criteria and scoring rules.

\section{\bench Construction Details}
\label{app:construction}

This appendix provides additional details on schema design, database population, domain rules, tool generation and validation, task synthesis, and quality control for \bench.

\subsection{Detailed Construction Pipeline}
\label{app:construction-pipeline}

The construction pipeline follows the three top-level stages shown in Figure~\ref{fig:ui_bench}, with each stage producing a validated artifact required by the next. During Tool Design, corresponding domain rules specify action requirements, service constraints, and valid state transitions before the tools are constructed and validated. This ordering prevents errors in collected data, rules, or tool behavior from propagating into the final benchmark instances. All pseudorandom choices in data construction use seed 42.

\paragraph{Stage 1: Domain Data Collection.} The benchmark uses 10 real-world domain databases constructed from the public data sources documented in Table~\ref{tab:domain-references}. The same annotators collect and preprocess these sources by standardizing field types and identifiers and checking cross-table references, then organize each domain into a user-facing entity table, resource tables, and a transaction table. The resulting databases are used for tool and task construction.

\begin{table}[t]
    \centering
    \caption{Public data sources for the 10 evaluation domains. The table summarizes the domain information provided by each source and its license or access terms.}
    \label{tab:domain-references}
    \footnotesize
    \setlength{\tabcolsep}{4pt}
    \renewcommand{\arraystretch}{1.12}
    \begin{tabular}{@{}>{\raggedright\arraybackslash}p{0.13\textwidth}>{\raggedright\arraybackslash}p{0.24\textwidth}>{\raggedright\arraybackslash}p{0.40\textwidth}>{\raggedright\arraybackslash}p{0.15\textwidth}@{}}
        \toprule
        \textbf{Domain} & \textbf{Public data source} & \textbf{Domain entities or workflows} & \textbf{License / terms} \\
        \midrule
        Flight booking & OpenFlights airport, airline, and route data~\citep{openflightsdata} & Airports, airlines, routes, equipment, and route-level connectivity & ODbL; component terms vary \\
        Hotel booking & Hotel Booking Demand datasets~\citep{antonio2019hotel} & Hotel bookings, arrival dates, room assignment, cancellation status, and special requests & CC BY 4.0 \\
        Bank account & Synthetic Mobile Money Transaction Dataset~\citep{azamuke2024mobilemoney} & Account identifiers, transfers, transaction amounts, and balance transitions & CC BY 4.0 \\
        Restaurant reservation & UCI Restaurant \& Consumer Data~\citep{medellin2011restaurant} & User profiles and preferences, restaurants, accepted payments, hours, and ratings & CC BY 4.0 \\
        E-commerce & Olist Brazilian E-Commerce Dataset~\citep{olist2018brazilian} & Customers, sellers, products, orders, payments, delivery status, and reviews & CC BY-NC-SA 4.0 \\
        Movie ticket & MovieLens 25M~\citep{harper2015movielens} & Pseudonymous users, movies, ratings, tags, and preference signals & GroupLens research-use terms \\
        Event ticketing & Ticketmaster Discovery API v2~\citep{ticketmasterdiscovery} & Events, venues, classifications, dates, locations, and availability filters & Ticketmaster developer terms \\
        Medical appointment & Medical Appointments No-Show~\citep{salazar2023medical} & Patient attributes, specialties, appointment time, attendance status, and no-show reasons & CC BY 4.0 \\
        Retail & UCI Online Retail II~\citep{chen2012onlineretailii} & Customers, products and SKUs, invoices, order lines, prices, and cancellations & CC BY 4.0 \\
        Car rental & Capacity-Pricing Model car-rental instances~\citep{oliveira2017capacitypricing} & Rental demand, vehicle groups, fleet capacity, and time- and price-dependent decisions & CC BY 4.0 \\
        \bottomrule
    \end{tabular}
\end{table}

\paragraph{Stage 2: Tool Design.} The same annotators first specify corresponding domain rules that govern action requirements, eligibility conditions, service constraints, and valid state transitions. They then define query and action tools over the normalized database. Query tools are atomic, read-only operations designed to compose into diverse search chains. Their combined coverage includes broad overview operations, such as listing the records available to a user-facing entity, and targeted retrieval operations, such as finding a resource by a date, location, status, or identifier. Action tools implement rule-compliant database writes across multiple service scenarios. Their database-valued arguments are grounded in actual query observations so that an agent can obtain the required values through the available tool interface. Each tool is paired with a schema that describes its purpose, arguments, constraints, and return values.

\paragraph{Tool Validation and Repair.} We validate query and action tools before using them for task synthesis. Each query tool is executed over the user-facing entity records, and each action tool is tested with inputs grounded in real query observations; action validation checks both compliance with the domain rules and the resulting database state change. An execution error or empty query result initiates a repair-and-revalidation loop. We re-execute the repaired implementation against the failed validation case and repeat this process at most twice. A candidate that remains unsuccessful after the second repair is dropped. We subsequently execute every proposed tool sequence end to end: a sequence is rejected if any tool raises an execution error or if it contains more than two empty query results. The surviving sequence must terminate in an executable reference action whose resulting database state exactly matches the target state. This separates artifact-level tool generation from task-level retention and ensures that only verified execution paths contribute to the benchmark.

\paragraph{Stage 3: Task Synthesis.} One of these annotators first samples a random user and specifies a service goal, such as changing an existing flight booking. Given the selected user's records, the domain database, and the available query tools, a task synthesis agent instantiated with \texttt{Claude Opus 5} proposes a chain of query calls for that goal. Each call draws its arguments from the database rather than from invented values, and later calls depend on the results of earlier ones, so that a chain traverses the schema the way an agent would, from the user record to an existing transaction and then to the resources that transaction involves. A chain contains 8.7 query calls on average. The synthesis agent executes the proposed chain against the domain database and then selects an action tool appropriate to the retrieved records. Each argument of that action is filled with a value observed during execution, so that every argument of the reference action is grounded in a retrieved record. We then remove information from the description the evaluated agent receives. Every internal identifier required by the action, such as a booking or record identifier, is masked, so the evaluated agent must retrieve it from the database. The user identifier remains visible so that the evaluated agent can determine which account the task concerns. Two ambiguity levels determine how many further arguments are masked. Moderate-ambiguity tasks mask 2 to 3 additional arguments selected at random and high-ambiguity tasks mask 4 to 5. Masked values are recoverable only through database exploration, user interaction, or both. Conditioned on the query chain, its execution results, and the domain rules, the synthesis agent generates the request the evaluated agent receives. The request states the user goal in natural language and leaves the masked arguments unspecified, so that the resulting ambiguity is what the evaluated agent must resolve. A candidate task is retained only if it passes three checks. An execution check re-executes the reference action and confirms that it produces the intended database state. An alignment check confirms that the user profile, the request, the retrieved records, the domain rules, and the reference action are mutually consistent. Two independent annotators confirm that the request is realistic for the domain and fluent and that no masked value appears; disagreements are resolved by a third annotator. Table~\ref{tab:bench-summary} reports inter-annotator agreement.

\subsection{Benchmark Statistics and Comparison}
\label{app:bench-stats}

Table~\ref{tab:domains} reports the database composition and task counts for each evaluation domain. The Lite and Full splits maintain comparable query length, ambiguity density, and annotation agreement, while Full broadens task and action-tool coverage (Table~\ref{tab:bench-summary}). Table~\ref{tab:benchmark-comparison} compares the interaction setting and evaluation target of closely related benchmarks. Tau-Bench and ToolSandbox elicit user information through dialogue. AppWorld evaluates generated programs that operate application APIs, while WebLINX evaluates agents that navigate existing websites under conversational instructions. In \bench, generated front-end code instead creates the interface through which the user provides task information; tool execution then connects that input to an objectively evaluated database update.

\begin{table}[htbp]
    \centering
    \caption{Comparison with related benchmarks, using the settings and scale reported in the cited papers. Tasks, scenarios, and demonstrations are distinct dataset units. Generated UI denotes front-end code produced for the user to operate, rather than code used by the agent to call APIs or navigate an existing interface.}
    \label{tab:benchmark-comparison}
    \small
    \setlength{\tabcolsep}{3.5pt}
    \renewcommand{\arraystretch}{1.10}
    \begin{tabular}{@{}>{\raggedright\arraybackslash}p{0.19\textwidth}>{\raggedright\arraybackslash}p{0.20\textwidth}>{\raggedright\arraybackslash}p{0.29\textwidth}>{\raggedright\arraybackslash}p{0.25\textwidth}@{}}
        \toprule
        \textbf{Benchmark} & \textbf{Reported scale} & \textbf{Interaction setting} & \textbf{Evaluation target} \\
        \midrule
        Tau-Bench~\citep{yao2024taubench} & 165 tasks; 2 domains & Agent converses with a simulated user and calls domain APIs & Final database state versus annotated goal \\
        \addlinespace[3pt]
        ToolSandbox~\citep{lu2025toolsandbox} & 1{,}032 scenarios; 34 tools & Agent converses with a simulated user and executes stateful tools & Intermediate and final milestones; forbidden events \\
        \addlinespace[3pt]
        AppWorld~\citep{trivedi2024appworld} & 750 tasks; 9 apps; 457 APIs & Agent writes and executes programs to operate app APIs & State-based unit tests for task completion and unintended changes \\
        \addlinespace[3pt]
        WebLINX~\citep{lu2024weblinx} & Approximately 2{,}300 demonstrations; over 150 websites & Agent follows dialogue instructions to navigate existing websites & Next-action prediction against expert demonstrations \\
        \addlinespace[3pt]
        \bench & 300 Lite and 1{,}000 Full tasks; 10 domains & Agent generates an executable UI for user input, then acts through tools & Final database-state exact match after UI interaction and action execution \\
        \bottomrule
    \end{tabular}
\end{table}

 \begin{table}[H]
 \centering
 \caption{\bench domain statistics. Every domain provides a user-facing entity table, resource tables, and a transaction table. Columns and records count fields and rows across these tables; the table also reports the retained tasks in each split.}
 \label{tab:domains}
 \small
 \setlength{\tabcolsep}{6pt}
 \renewcommand{\arraystretch}{1.08}
 \begin{tabular}{@{}llcccc@{}}
 \toprule
 \textbf{Domain} & \textbf{Domain-specific tables} & \textbf{Columns} & \textbf{Records} & \textbf{Lite} & \textbf{Full} \\
 \midrule
 Flight booking & flights, bookings, airports & 56 & 156 & 30 & 100 \\
 Hotel booking & hotels, room inventory, bookings & 58 & 70 & 30 & 100 \\
 Bank account & accounts, transactions, loans & 67 & 70 & 30 & 100 \\
 Restaurant reservation & restaurants, reservations, reviews & 63 & 70 & 30 & 100 \\
 E-commerce & products, orders, product reviews & 52 & 70 & 30 & 100 \\
 Movie ticket & movies, theaters, bookings & 60 & 70 & 30 & 100 \\
 Event ticketing & events, venues, ticket purchases & 63 & 70 & 30 & 100 \\
 Medical appointment & appointments, physicians, clinics & 65 & 70 & 30 & 100 \\
 Retail & products, orders, store locations & 44 & 70 & 30 & 100 \\
 Car rental & vehicles, branches, rental bookings & 48 & 70 & 30 & 100 \\
 \midrule
 \textbf{Total} & & \textbf{576} & \textbf{786} & \textbf{300} & \textbf{1,000} \\
 \bottomrule
 \end{tabular}
\end{table}

\paragraph{Tool Semantics Review.} Two of these annotators apply the validation criteria from Section~\ref{sec:bench-construction} to every surviving tool, not to a sample. Each tool is checked independently by a second annotator against the domain rules and the resulting database state change, and a tool is retained only when both annotators agree it is correct. A tool the two annotators cannot agree on, or whose behavior remains ambiguous after the repair loop described above, is discarded rather than retained.

\paragraph{Task Quality Review.} The same annotators apply this check to every candidate task that passes the execution and alignment checks from Section~\ref{sec:bench-construction}, again not to a sample. For each candidate, two annotators rate it independently, judging whether the request is realistic for the domain, fluent, and free of any masked value. A disagreement between these two initial annotators is resolved by a third annotator through discussion, and a candidate whose ambiguity this discussion cannot resolve is dropped rather than retained. Table~\ref{tab:bench-summary} reports agreement between the initial independent ratings separately for Lite (94.1\%) and Full (94.3\%).

\section{Training Details}
\label{app:training}

\subsection{Training Data Construction}
The training pool contains 40 domain databases, all disjoint from the 10 evaluation domains. Before optimization, we partition them into 25 SFT databases and 15 RL databases, with no database shared across the two sets. We use \texttt{DeepSeek V4 Flash} to generate end-to-end trajectories within \method-Harness and retain a trajectory only when its terminal action executes and reaches the target database state. Each retained SFT task contributes one interface-generation example for the GUI Coder Agent and one or more pre- or post-agent examples for the Tool Agent. The RL split supplies task prompts for GUI-Coder GRPO and Reward Auditor sampling; the SFT Tool Agent remains fixed during RL. Table~\ref{tab:training-data} summarizes the resulting data.

\begin{table}[h]
    \caption{Training-domain split and data volume (train/validation). Each task contributes one GUI-Coder example or RL prompt; an SFT task can contribute multiple Tool Agent examples.}
    \label{tab:training-data}
    \centering
    \small
    \setlength{\tabcolsep}{7pt}
    \renewcommand{\arraystretch}{1.05}
    \begin{tabular}{@{}lrrrr@{}}
        \toprule
        \textbf{Split} & \textbf{Databases} & \textbf{Tasks} & \textbf{Tool Agent data} & \textbf{GUI-Coder data} \\
        \midrule
        SFT & 25 & 12{,}376 / 254 & 32{,}409 / 669 & 12{,}376 / 254 \\
        RL  & 15 & 8{,}116 / 169  & --             & 8{,}116 / 169 \\
        \bottomrule
    \end{tabular}
\end{table}

\paragraph{ARIA snapshots and split verification.}
An ARIA snapshot records the rendered interface's accessible roles, names, and structure~\citep{w3c2023waiaria}. We use it as execution evidence for the reward judge. The 40 training databases and 10 evaluation domains are constructed from separate domain lists. We check for exact overlap in user-task text, serialized reference actions, and executable tool-name sequences; none is found. For recurring generic tool names, we additionally inspect arguments and implementations to check for reused tools or databases.

\subsection{Model and Optimization Details}

\paragraph{Models and role specialization.}
The Tool Agent and GUI Coder Agent use separate checkpoints initialized from \texttt{Qwen3.5-4B}~\citep{qwen2026qwen35_4b}. Both are first trained with full-parameter supervised fine tuning. In the \method-4B configurations, the Tool Agent is then kept fixed while only the GUI Coder Agent undergoes RL. Starting from the GUI Coder Agent SFT checkpoint, we train three independent RL policies for Direct RL, Audit Round~1, and Audit Round~2 (final). The final audited policy is GUI-Coder. The \method-4B designation refers to the active model size per call: although the system stores two role-specific checkpoints, each Tool Agent or GUI Coder Agent invocation activates only one 4B model. Table~\ref{tab:sft-config} summarizes the SFT settings.

\begin{table}[h]
    \caption{Supervised fine-tuning configuration for the two role-specific agents.}
    \label{tab:sft-config}
    \centering
    \footnotesize
    \renewcommand{\arraystretch}{0.95}
    \begin{tabular}{lcc}
        \toprule
        \textbf{Setting} & \textbf{Tool Agent} & \textbf{GUI Coder Agent} \\
        \midrule
        Training / validation examples & 32{,}409 / 669 & 12{,}376 / 254 \\
        Parameter update & Full fine-tuning & Full fine-tuning \\
        Optimizer & AdamW & AdamW \\
        Peak learning rate & $1\times10^{-5}$ & $1\times10^{-5}$ \\
        Weight decay / gradient clip & 0.05 / 1.0 & 0.05 / 1.0 \\
        Global batch size & 32 & 32 \\
        Maximum sequence length & 12{,}288 & 12{,}288 \\
        Schedule & \shortstack{Cosine; 20 warmup steps\\minimum LR $1\times10^{-6}$} & \shortstack{Cosine; 10 warmup steps\\minimum LR $1\times10^{-6}$} \\
        Configured epochs & 3 & 3 \\
        Data seed & 42 & 42 \\
        \bottomrule
    \end{tabular}
\end{table}

\paragraph{GRPO optimization.}
We optimize only the GUI Coder Agent with GRPO~\citep{shao2024deepseekmath}, keeping the Tool Agent fixed at its best SFT checkpoint. The reported sessions use 1{,}500 training and 41 validation prompts from the larger RL pool in Table~\ref{tab:training-data}. GRPO samples a group of interfaces for each prompt and uses their relative rewards to compute advantages, avoiding a separate value model. Let $\hat{A}_i$ denote the group-relative advantage of output $y_i$ and $\rho_{i,t}=\pi_\theta(y_{i,t}\mid x,y_{i,<t})/\pi_{\mathrm{old}}(y_{i,t}\mid x,y_{i,<t})$ the token-level importance ratio. The clipped policy objective, with the KL coefficient set to zero, is
\[
\mathcal{J}_{\mathrm{GRPO}}(\theta)=
\mathbb{E}\!\left[
\frac{1}{G}\sum_{i=1}^{G}\frac{1}{|y_i|}\sum_{t=1}^{|y_i|}
\min\!\left(
\rho_{i,t}\hat{A}_i,\;
\operatorname{clip}(\rho_{i,t},1-\epsilon,1+\epsilon)\hat{A}_i
\right)
\right],
\]
where $\epsilon$ controls the clipping range and $G$ is the number of outputs sampled per prompt. Table~\ref{tab:grpo-config} summarizes the shared training configuration.

\begin{table}[H]
    \caption{Shared GRPO configuration for the three independent GUI Coder Agent sessions.}
    \label{tab:grpo-config}
    \centering
    \footnotesize
    \renewcommand{\arraystretch}{0.95}
    \begin{tabular}{lp{0.67\linewidth}}
        \toprule
        \textbf{Setting} & \textbf{Value} \\
        \midrule
        Initialization & GUI Coder Agent SFT checkpoint \\
        Training / validation prompts & 1{,}500 / 41 \\
        Optimizer & AdamW; learning rate $5\times10^{-6}$; weight decay 0 \\
        Update budget & 500 updates; 32 rollouts per update \\
        Grouped sampling & 8 rollouts per prompt; oversampling factor 2.0 \\
        Sequence limit & 16{,}384 tokens \\
        Policy objective & GRPO clipped surrogate; KL coefficient 0 \\
        LoRA & Rank 16; alpha 32 \\
        Sampling & Temperature 1.0; top-$p$ 1.0; min-$p$ 0.05 \\
        Validation & Every 40 updates; 4 rollouts per validation prompt \\
        Seed & 42 \\
        Hardware & 8$\times$NVIDIA H200 GPUs \\
        Framework & \texttt{prime-rl} 0.7.0; \texttt{vLLM} 0.24.0; PyTorch 2.11.0+cu128 \\
        \bottomrule
    \end{tabular}
\end{table}

\subsection{Evaluation Inference Configuration}
\label{app:inference-config}

The open-weight baselines are \texttt{Qwen3.5-4B}, \texttt{Qwen3.6-35B-A3B}, and \texttt{Qwen3.6-27B}~\citep{qwen2026qwen35_4b,qwen2026qwen36_35b_a3b,qwen2026qwen36_27b}, \texttt{gpt-oss-120b}~\citep{openai2025gptoss}, \texttt{DeepSeek V4 Flash}~\citep{deepseekai2026v4flash0731}, and \texttt{Qwen3.5-397B-A17B}~\citep{qwen2026qwen35_397b_a17b}. The closed-source baselines are \texttt{GPT-5.4}~\citep{openai2026gpt54}, \texttt{Claude Sonnet 4.6}~\citep{anthropic2026sonnet46}, and \texttt{Claude Opus 5}~\citep{anthropic2026opus5}.

The Lite harness comparison in Table~\ref{tab:main} and the training ablation in Figure~\ref{fig:gui-coder-rl-progression} use temperature 0.7 for the evaluated model. All evaluation stages use a 32K-token output budget. Local \texttt{vLLM} requests leave top-$p$ unset and therefore use the serving default, whereas closed-source API requests and the user-simulator call explicitly use top-$p=1.0$. Open-weight models and our role-specific checkpoints are served through \texttt{vLLM}; the closed-source models are accessed through a hosted API under the aliases \texttt{openai-gpt-5.4}, \texttt{claude-sonnet-4-6}, and \texttt{claude-opus-5}. All three closed-source API baselines use the default thinking effort. The fixed user simulator is served as \texttt{DeepSeek-V4-Flash-0731}. These are the exact identifiers exposed to the evaluation pipeline; the serving APIs do not expose dated snapshots for these aliases.

Each structured generation permits at most three attempts when the required output cannot be extracted; transport-level API requests permit up to 30 attempts. These are internal retries within one evaluation session, distinct from the three complete sessions used for Pass@3 and Avg@3. Browser rendering and user-simulator execution each have a 60-second timeout. Each complete session starts from an isolated copy of the task's initial database state. Table~\ref{tab:main} reports all results to two decimal places.

\subsection{Bootstrap Confidence Intervals for Lite Results}
\label{app:lite-bootstrap}
To assess whether the aggregate advantages in Table~\ref{tab:main} are robust to task sampling, we perform a paired bootstrap over the 300 Lite tasks. Using random seed 42, in each of 10{,}000 replicates we sample task identifiers with replacement, retain all three sessions for every sampled task, and apply the same sampled identifiers to every harness and backbone. We compute the paired difference for each backbone and then macro-average across the fixed set of nine evaluated backbones. Table~\ref{tab:lite-bootstrap} reports percentile 95\% confidence intervals. All intervals remain above zero for both metrics, showing that the aggregate gains of \method-Harness are consistent across bootstrap resamples.

\begin{table}[H]
    \centering
    \caption{Macro-average paired differences across the nine shared backbones on Lite. Intervals are 95\% task-paired bootstrap confidence intervals over 10{,}000 resamples.}
    \label{tab:lite-bootstrap}
    \small
    \setlength{\tabcolsep}{12pt}
    \renewcommand{\arraystretch}{1.08}
    \begin{tabular}{@{}lcc@{}}
        \toprule
        Baseline & $\Delta$ Pass@3 (pp), 95\% CI & $\Delta$ Avg@3 (pp), 95\% CI \\
        \midrule
        \texttt{smolagents}     & +4.48 [2.37, 6.59] & +3.56 [2.15, 5.00] \\
        \texttt{mini-swe-agent} & +5.81 [3.37, 8.26] & +3.00 [1.46, 4.58] \\
        \texttt{Pi}             & +7.15 [4.30, 9.96] & +4.04 [2.30, 5.81] \\
        \bottomrule
    \end{tabular}
\end{table}

\subsection{Baseline Harness Adaptation}
\label{app:harness-adaptation}

The three baselines retain their original agent loops and text-based user interaction; we do not insert the GUI Coder Agent or renderer into them. They share the benchmark tasks, database, Python implementations of the query and action tools, user-simulator model and hidden goal, and final-state evaluator. When a baseline requests information, the simulator receives its textual message and interaction history and replies in natural language. This gives an end-to-end text-only comparison while keeping the task environment and success criterion fixed.

We adapt only the tool and termination boundaries required by each framework. In \texttt{smolagents}, the benchmark tools are registered through its native agent API. In \texttt{mini-swe-agent}, we replace the default shell-only action surface with the benchmark's JSON-schema tool set and add an explicit finish action. In \texttt{Pi}, its \texttt{pi-agent-core} Node.js loop accesses the same Python tools through a localhost HTTP bridge. Where a framework does not natively impose the benchmark's termination conditions, an external check applies the configured stopping rule.

Some role-specific checkpoints emit tagged tool calls rather than native function calls. For these checkpoints, a format adapter renders the expected prompt and maps tagged outputs into each framework's internal tool-call representation. This conversion does not modify generated arguments, tool observations, database state, or grading; beyond these compatibility layers, we retain each framework's own agent loop.

\subsection{Full Split Results}
\label{app:full-results}

The Full split contains 1{,}000 tasks, each requiring end-to-end tool use, user interaction, and state evaluation; \method-Harness additionally generates and executes an interface. Due to the resulting computational cost, we evaluate each task in one session and restrict this four-harness comparison to open-weight models. Table~\ref{tab:full-results} reports single-session Success Rate, equivalently Pass@1. Under the same Full protocol, our role-specialized \method-4B achieves \textbf{40.90\%} Success Rate.

\begin{table}[H]
    \centering
    \caption{State EM Pass@1 (single-session Success Rate, \%) on the Full split of \bench (1{,}000 tasks) across four agent harnesses in their native interaction modes. Bold marks the best result in each row.}
    \label{tab:full-results}
    \small
    \setlength{\tabcolsep}{9pt}
    \renewcommand{\arraystretch}{1.08}
    \begin{tabular}{@{}lcccc@{}}
        \toprule
        \textbf{Model} & \cellcolor{citecolor!7}\textbf{\method-Harness} & \textbf{smolagents} & \textbf{mini-swe-agent} & \textbf{Pi} \\
        \midrule
        \rowcolor{black!4}
        \multicolumn{5}{@{}l}{\hspace{0.45em}\textsc{Open-weight models}} \\
        \addlinespace[1pt]
        \texttt{Qwen3.5-4B}        & \cellcolor{citecolor!7} 4.11 &  3.80 &  4.70 & \textbf{5.30} \\
        \texttt{Qwen3.6-27B}       & \cellcolor{citecolor!7}\textbf{37.50} & 34.50 & 21.00 & 27.00 \\
        \texttt{Qwen3.6-35B-A3B}   & \cellcolor{citecolor!7}\textbf{35.30} & 27.10 & 24.50 & 21.90 \\
        \texttt{gpt-oss-120b}      & \cellcolor{citecolor!7}\textbf{32.20} & 14.50 & 30.00 & 25.50 \\
        \texttt{DeepSeek V4 Flash} & \cellcolor{citecolor!7}\textbf{33.00} & 31.80 & 29.40 & 28.80 \\
        \texttt{Qwen3.5-397B-A17B} & \cellcolor{citecolor!7}\textbf{37.70} & 36.80 & 28.00 & 29.70 \\
        \bottomrule
    \end{tabular}
\end{table}

\subsection{Dynamic UX Systems Benchmark}
\label{app:dynamic-ux}
We compare two process-management strategies on 1{,}100 pre-verified successful TSX components, sampling 100 from each of 11 evaluated model configurations with seed 42. The fresh-process implementation launches and tears down a Chromium browser for every rollout. Dynamic UX launches one persistent browser instance and creates a fresh, isolated \texttt{browser.new\_context()} for every rollout. Both use the same per-render identifier, websocket messages, component-loaded signal, and dedicated websocket-server--Vite pair; only the browser lifecycle differs.

We evaluate both implementations at concurrency 1, 4, 8, 16, and 32 in the same 16-core Rocky Linux 9.8 sandbox. Each configuration is repeated three times with newly shuffled sample orders drawn from a pseudorandom stream initialized with seed 42. The first \emph{concurrency} completed requests in each session are treated as warm-up and excluded from latency and throughput, but remain in failure and timeout counts. A 200-ms background sampler records RSS and the number of Chromium processes in the harness process tree. Table~\ref{tab:dynamic-ux-systems} averages throughput and latency across repeats; peak RSS and process count are the maxima observed across the three repeats.

\begin{table}[h]
    \caption{Matched-concurrency rendering benchmark. Baseline launches one browser per rollout; Dynamic UX shares one browser across isolated contexts. Bold marks the better value; green arrows show relative change from Baseline.}
    \label{tab:dynamic-ux-systems}
    \centering
    \footnotesize
    \setlength{\tabcolsep}{4.5pt}
    \renewcommand{\arraystretch}{1.00}
    \begin{tabular*}{\textwidth}{@{\extracolsep{\fill}}c l r r r r r@{}}
        \toprule
        \textbf{C} & \textbf{System} & \shortstack{\textbf{Renders/min}\\$\uparrow$} & \shortstack{\textbf{Median (ms)}\\$\downarrow$} & \shortstack{\textbf{P95 (ms)}\\$\downarrow$} & \shortstack{\textbf{Peak RSS (MB)}\\$\downarrow$} & \shortstack{\textbf{Chromium processes}\\$\downarrow$} \\
        \midrule
        \multirow{2}{*}{\raisebox{-0.9\baselineskip}{1}}  & Baseline & 25.48 & 2{,}237.5 & 2{,}349.1 & 894.7 & 7 \\
        & \cellcolor{rewardauditbg}\textbf{Dynamic UX} & \cellcolor{rewardauditbg}\shortstack{\textbf{26.47}\\\textcolor{green!50!black}{\scriptsize($\uparrow$3.89\%)}} & \cellcolor{rewardauditbg}\shortstack{\textbf{2{,}226.5}\\\textcolor{green!50!black}{\scriptsize($\downarrow$0.49\%)}} & \cellcolor{rewardauditbg}\shortstack{\textbf{2{,}328.8}\\\textcolor{green!50!black}{\scriptsize($\downarrow$0.86\%)}} & \cellcolor{rewardauditbg}\shortstack{\textbf{876.1}\\\textcolor{green!50!black}{\scriptsize($\downarrow$2.08\%)}} & \cellcolor{rewardauditbg}\shortstack{\textbf{6}\\\textcolor{green!50!black}{\scriptsize($\downarrow$14.29\%)}} \\
        \addlinespace[1pt]
        \multirow{2}{*}{\raisebox{-0.9\baselineskip}{4}}  & Baseline & 94.30 & 2{,}471.8 & 2{,}863.4 & 2{,}306.8 & 24 \\
        & \cellcolor{rewardauditbg}\textbf{Dynamic UX} & \cellcolor{rewardauditbg}\shortstack{\textbf{101.23}\\\textcolor{green!50!black}{\scriptsize($\uparrow$7.35\%)}} & \cellcolor{rewardauditbg}\shortstack{\textbf{2{,}271.2}\\\textcolor{green!50!black}{\scriptsize($\downarrow$8.12\%)}} & \cellcolor{rewardauditbg}\shortstack{\textbf{2{,}382.5}\\\textcolor{green!50!black}{\scriptsize($\downarrow$16.79\%)}} & \cellcolor{rewardauditbg}\shortstack{\textbf{1{,}556.3}\\\textcolor{green!50!black}{\scriptsize($\downarrow$32.53\%)}} & \cellcolor{rewardauditbg}\shortstack{\textbf{10}\\\textcolor{green!50!black}{\scriptsize($\downarrow$58.33\%)}} \\
        \addlinespace[1pt]
        \multirow{2}{*}{\raisebox{-0.9\baselineskip}{8}}  & Baseline & 144.18 & 3{,}188.8 & 4{,}076.9 & 4{,}388.7 & 48 \\
        & \cellcolor{rewardauditbg}\textbf{Dynamic UX} & \cellcolor{rewardauditbg}\shortstack{\textbf{189.03}\\\textcolor{green!50!black}{\scriptsize($\uparrow$31.11\%)}} & \cellcolor{rewardauditbg}\shortstack{\textbf{2{,}410.1}\\\textcolor{green!50!black}{\scriptsize($\downarrow$24.42\%)}} & \cellcolor{rewardauditbg}\shortstack{\textbf{2{,}548.1}\\\textcolor{green!50!black}{\scriptsize($\downarrow$37.50\%)}} & \cellcolor{rewardauditbg}\shortstack{\textbf{2{,}525.0}\\\textcolor{green!50!black}{\scriptsize($\downarrow$42.47\%)}} & \cellcolor{rewardauditbg}\shortstack{\textbf{15}\\\textcolor{green!50!black}{\scriptsize($\downarrow$68.75\%)}} \\
        \addlinespace[1pt]
        \multirow{2}{*}{\raisebox{-0.9\baselineskip}{16}} & Baseline & 188.11 & 4{,}870.2 & 6{,}467.4 & 8{,}158.7 & 96 \\
        & \cellcolor{rewardauditbg}\textbf{Dynamic UX} & \cellcolor{rewardauditbg}\shortstack{\textbf{258.64}\\\textcolor{green!50!black}{\scriptsize($\uparrow$37.49\%)}} & \cellcolor{rewardauditbg}\shortstack{\textbf{3{,}445.7}\\\textcolor{green!50!black}{\scriptsize($\downarrow$29.25\%)}} & \cellcolor{rewardauditbg}\shortstack{\textbf{3{,}798.8}\\\textcolor{green!50!black}{\scriptsize($\downarrow$41.26\%)}} & \cellcolor{rewardauditbg}\shortstack{\textbf{4{,}235.0}\\\textcolor{green!50!black}{\scriptsize($\downarrow$48.09\%)}} & \cellcolor{rewardauditbg}\shortstack{\textbf{23}\\\textcolor{green!50!black}{\scriptsize($\downarrow$76.04\%)}} \\
        \addlinespace[1pt]
        \multirow{2}{*}{\raisebox{-0.9\baselineskip}{32}} & Baseline & 214.08 & 8{,}724.7 & 10{,}218.6 & 15{,}440.0 & 192 \\
        & \cellcolor{rewardauditbg}\textbf{Dynamic UX} & \cellcolor{rewardauditbg}\shortstack{\textbf{238.32}\\\textcolor{green!50!black}{\scriptsize($\uparrow$11.32\%)}} & \cellcolor{rewardauditbg}\shortstack{\textbf{7{,}467.2}\\\textcolor{green!50!black}{\scriptsize($\downarrow$14.41\%)}} & \cellcolor{rewardauditbg}\shortstack{\textbf{8{,}846.3}\\\textcolor{green!50!black}{\scriptsize($\downarrow$13.43\%)}} & \cellcolor{rewardauditbg}\shortstack{\textbf{7{,}585.8}\\\textcolor{green!50!black}{\scriptsize($\downarrow$50.87\%)}} & \cellcolor{rewardauditbg}\shortstack{\textbf{39}\\\textcolor{green!50!black}{\scriptsize($\downarrow$79.69\%)}} \\
        \bottomrule
    \end{tabular*}
\end{table}

At concurrency 1, the two architectures perform similarly. As concurrency increases, repeatedly launching browser process trees becomes increasingly expensive. Dynamic UX reaches its highest throughput at concurrency 16, where it improves throughput by 37.5\%, reduces median and P95 latency by 29.2\% and 41.3\%, reduces peak RSS by 48.1\%, and uses 76.0\% fewer Chromium processes than fresh-process rendering. At concurrency 32, both architectures approach saturation; Dynamic UX still provides 11.3\% higher throughput with 50.9\% lower peak RSS and 79.7\% fewer Chromium processes. Separately, we render 1{,}000 sessions containing unique marker strings through Dynamic UX at concurrency 32. All sessions complete without timeout or rendering failure, and no page contains another session's marker.

\subsection{Initial Reward Specification}
\label{app:initial-reward}

The initial rubric translates the requirements of a task-completing interface into four criteria. Functional equivalence assesses whether the candidate supports the task-relevant functions of the successful reference; feature completeness checks whether the required information and controls are present; implementation correctness concerns the generated code and component behavior; and task alignment checks consistency with the user request. The reference is drawn from a successful trajectory, providing a concrete example of task completion rather than a code string that the candidate must reproduce. The online judge, \texttt{DeepSeek V4 Flash} at temperature 0, receives the query, reference and candidate code, and their initial rendered ARIA snapshots. It returns scores on these four criteria and supporting rationales. Section~\ref{sec:reward-auditor} gives the normalized weighted score.

Direct RL uses this initial judge score after TSX extraction, task/reference-evidence checks, structural validation, rendering, and valid judge-response checks. A failed check receives zero reward. Reference ARIA is cached before training, whereas candidate code is rendered for each rollout. This online path does not execute the full user-interaction and post-agent trajectory for every rollout; those executions provide the objective evidence used by Reward Auditor. Collecting one such outcome requires browser interaction, post-agent inference, action execution, and terminal-state evaluation. Reward Auditor applies this evidence to a sampled audit panel and amortizes the limited execution budget by converting its diagnostics into a shared reward specification used across RL rollouts.

\subsection{Meta-Reward Revision Protocol}
\label{app:reward-design}

\paragraph{Execution-based diagnosis.} Reward Auditor diagnoses and revises rewards from internal rollouts on the 15 RL training databases. \texttt{DeepSeek V4 Flash}~\citep{xu2026deepseek} generates a Playwright program to interact with the rendered interface; after browser execution, the fixed Tool Agent consumes the submission and predicts the terminal action. The objective outcome is positive only when the action executes and the resulting database state exactly matches the target. Reward Auditor compares this outcome with $\hat{z}_i=\mathbb{I}[r_i\geq0.5]$, where $r_i$ is the complete reward, not the raw judge score. The auditor examines the resulting true/false positive and negative cases, aggregate reward distributions, and complete trajectories, including retrieved evidence, generated code, user submissions, final actions, and judge rationales, to diagnose failure causes and recurring errors in the evaluation criteria or score assignment.

\paragraph{Automated revision, review, and replay.} In each round, Reward Auditor modifies the current reward specification from reward-distribution summaries, diagnosed error patterns, and individual cases, and reviews the proposed changes. It replays candidate specifications on the same stored outputs and reviews their effects on successful and unsuccessful traces before accepting a revision. Diagnosis, revision, and review require no human intervention. Holding the generated outputs fixed isolates score changes due to the reward specification. Development checks include replay of 1{,}618 Direct RL rollouts from 372 groups for the deterministic scoring rules and comparison of candidate judges on 1{,}005 paired frozen candidates.

\paragraph{Iterative reward revision.} Audit Round~1 and Audit Round~2 are successive meta-reward-guided revisions of the training feedback. The revisions address both what the judge evaluates and how its judgments and execution evidence determine the final scalar. Table~\ref{tab:reward-repairs} summarizes the targeted problems. After automated review and replay, task-specific diagnoses are consolidated into one shared rubric and scoring specification. The accepted reward is applied across tasks and remains fixed for its RL session. Direct RL and both audited variants train independent policies from the same GUI Coder Agent SFT checkpoint, with the Tool Agent fixed; reward versions are iterated, not policy checkpoints continued along one training trajectory.

\begin{table}[t]
    \caption{Problems targeted by meta-reward-guided reward revision. Revisions combine judging criteria with deterministic scoring rules.}
    \label{tab:reward-repairs}
    \centering
    \footnotesize
    \setlength{\tabcolsep}{5pt}
    \renewcommand{\arraystretch}{1.08}
    \begin{tabular}{@{}>{\raggedright\arraybackslash}p{0.29\textwidth}>{\raggedright\arraybackslash}p{0.36\textwidth}>{\raggedright\arraybackslash}p{0.29\textwidth}@{}}
        \toprule
        \textbf{Diagnosed problem} & \textbf{Reward revision} & \textbf{Intended effect} \\
        \midrule
        Plausible interfaces receive high scores despite interaction or payload defects & Revise judging criteria for actionable controls, submission wiring, payload semantics, and task fidelity & Make task-blocking defects explicit in the reward \\
        Different failure stages collapse to the same zero score & Assign stage-aware low-reward bands & Distinguish progress within unsuccessful rollouts \\
        Generated code uses unavailable icon exports & Apply a deterministic unsupported-icon gate & Prevent the quality score from overriding the detected defect \\
        Label forms or associations impede control access & Apply source-based accessibility penalties & Include interaction-relevant defects in score assignment \\
        \bottomrule
    \end{tabular}
\end{table}

\paragraph{Final reward specification.}
The final reward combines the four-dimensional quality score with ordered failure checks and a source-based accessibility multiplier. Using the judge scores defined in Section~\ref{sec:reward-auditor}, we compute
\[
Q_i=\frac{0.35q_i^{\mathrm{fe}}+0.35q_i^{\mathrm{fc}}+0.15q_i^{\mathrm{ic}}+0.15q_i^{\mathrm{ta}}}{5}.
\]
This uses the initial quality-score weights without additional rescaling. For each soft check $c$, let $s_{i,c}\in[0,1]$ be the fraction of eligible controls or labels exhibiting that defect. The accessibility penalty and multiplier are
\[
P_i=\frac{0.110s_{i,\mathrm{plain}}+0.055s_{i,\mathrm{long}}+0.056s_{i,\mathrm{link}}}{0.221},
\qquad A_i=1-0.5\min(1,P_i).
\]
The three nonzero-weight checks detect labels that are not plain strings, overlong labels (more than 5 words or 39 characters), and \texttt{Select} controls without a \texttt{labelId} linked to an \texttt{InputLabel}. Their weights are 0.110, 0.055, and 0.056, respectively; the label-collision check has weight zero and does not change $P_i$. Severities are proportions rather than binary flags: three defective labels among nine eligible labels give severity $1/3$. These checks form one shared scoring specification developed through the internal reward-audit process.

The following rules are evaluated from top to bottom, and the first matching branch returns the reward:
\[
r_i^{(2)}=
\begin{cases}
0.00, & \text{masked rollout},\\
0.00, & \text{TSX extraction fails},\\
0.02, & \text{format check fails},\\
0.05, & \text{code is incomplete},\\
0.05+0.05(1-s_{i,\mathrm{icon}}), & \text{an unavailable icon is used},\\
0.10, & \text{rendering fails},\\
0.15, & \text{rendering succeeds but ARIA has fewer than 400 characters},\\
0.20+0.80Q_iA_i, & \text{otherwise}.
\end{cases}
\]
A rollout is masked if the judge reports an error, rendering reports a transient error, or reference evidence (including the reference ARIA snapshot) is missing. Here $s_{i,\mathrm{icon}}$ measures the proportion of icon imports that are used but unavailable in the component package. The icon check is a gate and does not enter the soft accessibility penalty or proceed to quality scoring. Reported transient rendering errors are masked at zero; the 0.10 branch applies to unmasked rendering failures. The short-ARIA branch is a low-content heuristic. The multiplier satisfies $A_i\in[0.5,1]$, so the quality-scoring branch lies in $[0.20,1.00]$, above all failure-stage scores. For example, a quality score of 0.80 with no accessibility penalty gives 0.84; a perfect quality score with all three soft-check severities equal to one gives 0.60. In contrast, the initial reward assigns zero to every failed check and otherwise returns $Q_i$.

\paragraph{Intermediate reward revision.}
Audit Round~1 introduced the judge caps in Table~\ref{tab:audit-round-one-caps}. It is an intermediate reward version; the final reward is the ordered specification above.

\begin{table}[H]
    \caption{Explicit judge caps in the Audit Round~1 revision. FE, FC, IC, and TA denote functional equivalence, feature completeness, implementation correctness, and task alignment.}
    \label{tab:audit-round-one-caps}
    \centering
    \scriptsize
    \setlength{\tabcolsep}{5pt}
    \renewcommand{\arraystretch}{1.08}
    \begin{tabular}{@{}>{\raggedright\arraybackslash}p{0.42\textwidth}>{\raggedright\arraybackslash}p{0.52\textwidth}@{}}
        \toprule
        \textbf{Detected defect} & \textbf{Maximum judge score} \\
        \midrule
        Empty/error interface or no actionable control & All four dimensions $=0$ \\
        Missing or unreachable \texttt{submitData} & FE, TA $\leq 1$ \\
        Missing required controls & FC $\leq 4,3,2$ when one, two, or at least half are missing \\
        Wrong, constant, or omitted submitted values & FE, TA $\leq 3$ \\
        Broken control wiring & IC $\leq 3$; TA $\leq 3$ when task-blocking \\
        Harmful invented facts & TA $\leq 2$ \\
        Contradicted task evidence or required-language violation & TA $\leq 3$ \\
        \bottomrule
    \end{tabular}
\end{table}

The Audit Round~1 parser also lowers an unsupported perfect score from 5 to 4.5 and applies deterministic dimension caps to nonempty critical, payload, interaction, and visible-content defect lists. These rules illustrate how diagnosed defects become explicit evaluation criteria. Subsequent revision is guided by the remaining reward--outcome mismatches, with both false positives and false negatives considered rather than requiring every criterion to become stricter.

\section{Detailed Error Analysis}
\label{app:error-analysis}

\paragraph{Evaluation cohort.}
We analyze one session on each of the 300 Lite tasks for \texttt{DeepSeek V4 Flash}, \texttt{Claude Sonnet 4.6}, \texttt{Claude Opus 5}, and \method-4B, giving 1{,}200 model-task sessions. These single-session diagnostics are separate from the three-session aggregates in Table~\ref{tab:main}. Table~\ref{tab:error-stage} partitions the recorded outcomes. Across 800 failures, 416 (52.0\%) have no valid UI interaction, 356 (44.5\%) are classified as correct-tool calls with mismatched arguments or resulting state, and 28 (3.5\%) are rejected during execution.

\begin{table}[H]
    \caption{Single-session outcome partition on the 300 Lite tasks under \method-Harness. Columns are mutually exclusive and sum to 300 for every configuration.}
    \label{tab:error-stage}
    \centering
    \footnotesize
    \setlength{\tabcolsep}{6pt}
    \renewcommand{\arraystretch}{1.05}
    \begin{tabular}{@{}lrrrrr@{}}
        \toprule
        \textbf{Configuration} & \textbf{No valid interact.} & \textbf{Wrong tool} & \textbf{Wrong args} & \textbf{Exec.\ rejected} & \textbf{Success} \\
        \midrule
        \texttt{DeepSeek V4 Flash} & 124 & 0 & 71  & 7  & 98 \\
        \texttt{Claude Sonnet 4.6} & 87  & 0 & 102 & 10 & 101 \\
        \texttt{Claude Opus 5}     & 82  & 0 & 124 & 1  & 93 \\
        \textbf{\method-4B (Ours)} & 123 & 0 & 59  & 10 & \textbf{108} \\
        \bottomrule
    \end{tabular}
\end{table}

\paragraph{Diagnostic evidence.}
We inspect generated TSX, rendered ARIA snapshots, user-interaction traces, submitted payloads, and final actions to distinguish failure stages from underlying causes. For detailed trace review, we sample eight no-valid-interaction failures and eight argument-mismatch failures per model (64 cases in total; seed 42). These cases support the mechanisms summarized in Table~\ref{tab:error-mechanisms}; their frequencies are not extrapolated to all failures. The ``no valid interaction'' category records where execution stops, not whether the underlying cause is the interface or the simulator. We attribute a failure to generated code when the trace exposes a concrete restriction or incorrect binding that prevents the required interaction or submission.

\begin{table}[htbp]
\centering
\caption{Observed failure mechanisms and the evidence used to identify them.}
\label{tab:error-mechanisms}
\small
\begin{tabular}{@{}>{\raggedright\arraybackslash}p{0.25\textwidth}>{\raggedright\arraybackslash}p{0.69\textwidth}@{}}
\toprule
\textbf{Mechanism} & \textbf{Trace evidence} \\
\midrule
Render failure & Compiler or runtime errors prevent an interactive page from loading. \\
Inaccessible control & A control's rendered accessible name or label binding does not support the attempted browser interaction. \\
Insufficient input controls & The available controls cannot express the required selection or value. \\
Incorrect submitted value & The payload diverges from the task requirement or displayed choice, with no control to correct it. \\
\bottomrule
\end{tabular}
\end{table}

\paragraph{Representative cases.}
In a hotel-preference task, a single-select radio group cannot represent the requested selection of two hotels. In a flight-booking task, an immutable cabin-class field submits ``Economy'' although the task requires ``Premium Economy.'' These restrictions arise from the generated controls and submission logic: knowing the intended selection is insufficient when the UI cannot represent or submit it. The examples show how generated code can obstruct interaction even when the user's goal is fully specified.

\paragraph{Interpreting state mismatches.}
Trace inspection also finds differences involving generated identifiers, list ordering, and free-text values. These remain failures under the reported strict database-state comparison. A state mismatch alone does not establish whether its cause lies in the UI payload, downstream tool reasoning, or state-comparison behavior; that attribution requires the corresponding execution trace and domain semantics.

\section{Communication-Channel Survey Protocol}
\label{app:human-study}

\paragraph{Survey design and task selection.}
We compare text-based and generated-UI communication on the same task content to assess interaction effort and perceived latency. This comparison focuses on how reviewers communicate their preferences once an executable interface is available. The survey involves six human reviewers and 100 tasks from \bench Lite. All reviewers provided informed consent. Reviewers were compensated with gift cards at a rate of US\$30 per hour. Participation is anonymous, and we collect no names or other identifying information. All user profiles and task records shown in the survey are synthetic, and the tasks concern routine service workflows such as booking and reservation changes rather than sensitive personal decisions. Reviewers only complete the assigned interaction and assess its usability or latency; no real customer data or private information is involved. Interface content comes from four frontier models: \texttt{GPT-5.4}, \texttt{Claude Opus 5}, \texttt{Claude Sonnet 4.6}, and \texttt{DeepSeek V4 Flash}. For each model, a task must execute the generated interface successfully and produce a valid user-simulator interaction, including a usable submission, in all three sessions. We intersect the four models' eligible task sets, sample 100 tasks from this common pool, and fix one source model for each selected task. Let $v_{i,m,j}=1$ when session $j$ of model $m$ on task $i$ meets both validity conditions and $0$ otherwise. With $\mathcal{M}$ denoting the four source models, eligibility requires
\[
\prod_{m\in\mathcal{M}}\prod_{j=1}^{3}v_{i,m,j}=1.
\]
We denote the all-three requirement for each model by $\mathrm{pass}^{3}$: the superscript indicates that every session must pass the interface-validity checks. This differs from Pass@3, which requires at least one successful session. Here, validity concerns interface execution and interaction, rather than final database-state exact match. The common eligibility rule keeps the channel comparison from being confounded by rendering or submission failures and applies equally to all four content sources. The resulting comparison concerns communication on consistently usable interfaces; benchmark state-EM results separately include unsuccessful executions and interactions.

\paragraph{Assignment and paired conditions.}
For each task, the six reviewers are randomly assigned with seed 42 to two equal groups: three interact through text and three through the generated UI. Each reviewer encounters every task in only one condition, giving 300 task--reviewer interactions per condition. This design pairs conditions at the task level and prevents reviewers from carrying answers learned in one version into the other version of the same task.

Both conditions use the same user query, retrieved options, and task state from one fixed source model per task. In the UI condition, reviewers inspect the interface and provide or revise values through its controls. A fixed \texttt{DeepSeek V4 Flash} converter produces the text condition from the corresponding interface's ARIA snapshot, preserving concrete values, prefilled preferences, and unresolved fields; reviewers respond by typing. The conversion holds the available task information fixed so that the comparison concerns its presentation through text or UI, rather than differences in independently retrieved content. Neither condition exposes the hidden reference action or internal execution decisions.

\paragraph{Real-user interaction and stopping criterion.}
Reviewers can ask follow-up questions or request confirmation in either condition, with responses provided by our trained agent. An interaction ends when the agent successfully executes an action that produces the target database state. This evaluation permits the exchanges needed to complete the task rather than imposing the benchmark's fixed pre-agent--UI--post-agent sequence.

\paragraph{Interaction rounds.}
One round consists of a user message or interface submission and the corresponding agent response. Multiple field edits or clicks within one UI submission do not count as separate dialogue rounds. We count exchanges through successful task completion and average across the 300 task--reviewer interactions in each condition. The resulting means are 3.4 rounds for text and 1.2 for generated UI.

\paragraph{UI latency and acceptability.}
Interfaces are generated live during the survey. UI latency is measured from the start of interface generation through rendering and display. The three reviewers assigned to the UI condition independently assess latency acceptability for that task. A task passes when at least two of them judge its latency acceptable; the reported rate is the number of passing tasks divided by 100. This criterion is met for 85/100 tasks (85\%). These outcomes describe the selected tasks with consistently valid interfaces.

\section{Limitations}
\label{app:limitations}

Benchmark evaluation uses an LLM-based user simulator operating through Playwright. The simulator receives generated source and the reference goal to construct browser actions; although it must act through the rendered interface and cannot invoke task actions directly, this information may make it more capable than a user who observes only the interface. Source-blind evaluation from human-observable state is an important direction for future work. The reviewer survey covers a selected subset of consistently usable interfaces. Code execution and accessibility-tree evidence may not fully capture visual usability, including overlapping layouts or confusing arrangements. Strict database-state equality can also reject semantically equivalent outcomes that differ in generated identifiers, ordering, or free text. Adaptive interface regeneration after further clarification is an important direction for future work.

\input{prompt}
\end{document}

%% file: prompt.tex
\section{Prompt Templates}
\label{app:prompts}

This appendix gives the prompt templates for the agent and simulator roles in \method-Harness, lightly
copy-edited for readability. The Tool Agent uses two templates, one for pre-agent exploration and one
for post-agent execution, matching the two stages of Section~\ref{sec:genui-harness}. Fields written as
\texttt{[[name]]} are placeholders substituted before each call. \texttt{[[agent\_history]]} carries
the running transcript of thought, action, and observation triples, \texttt{[[domain\_db]]} and
\texttt{[[tool\_sets]]} the sampled domain records and available tool schemas, and
\texttt{[[reasoning\_for\_ui]]} the output of an intermediate reflection step that selects which
retrieved values the interface must expose. Lines exceeding the text width are wrapped for printing. We additionally give the initial four-criterion reward judge template. Appendix~\ref{app:reward-design} specifies the final reward composition, including ordered failure checks and accessibility penalties. The meta-reward
revision step uses \texttt{Claude Opus 5} to analyze reward distributions and individual errors,
revise and review the criteria, and assess replay results automatically. Appendix~\ref{app:reward-design}
describes this revision protocol.

\begin{tcolorbox}[breakable, title=Tool Agent pre-agent exploration template]
\begin{lstlisting}[basicstyle=\scriptsize\ttfamily, breaklines=true, columns=fullflexible]
You are a helpful tool agent. You are good at performing tasks with tools.
You will be given a task, your interaction history with the environment, and a set of tools. Think step by step about how to use the query tools to retrieve the information from the domain database that you need in order to call an action tool for the user.
You have only one chance to ask the user all the questions needed to complete the task.
First think about what you need to do, then decide which tool to use, then perform the action by writing its arguments.

Specifically, you should:
1. Think about the task and what is needed to complete it, and fill in the Thought field.
2. Decide which query tool to use and fill in the Action field with the tool name and the arguments that do not come from the databases.
3. Use this information-gathering step to prepare better clarification questions for the user.

# Notice:
- Always think step by step, and always use the tools to perform the action.
- Always fill in the Thought and Action fields, and never skip a step.
- Perform one step at a time, and wrap each tool call in <tool></tool> tags.
- Because of permission constraints, you may only use query-type tools on the domain database.
- To save space, do NOT mention user_database or domain_database in the Action field. Doing so breaks the program.
- The tool in the Action field must be given in full, with nothing omitted except database contents.
- You must NOT use tools outside the tool sets provided in the Task field. Doing so causes an error.
- Do not hallucinate arguments. Take them from the sample databases and from other tool responses.
- Only query tools are needed at this phase. Action tools are shown so you can see which arguments they require.
- You must wrap your thought in <thought></thought> tags and your response in ```agent_state ... ```.

You will be given a task, a set of tools, and the relevant databases. The task is described in the Task field.

# Here are sample domain and user databases for precise tool calling:
## Sample domain database:
[[domain_db]]

# Tool sets:
[[tool_sets]]

# User Id: [[user_id]]
# User Task: [[task]]

# Agent History:
[[agent_history]]

Your response must be in the following format, with all tags kept:

```agent_state
# Step [[step_num]]:
Thought:
<thought>// Your thought here</thought>
Action:
<tool>
// Your tool call here. One example is:
{
 "type": "query",
 "function": {
 "name": "get_comic_arc_info",
 "arguments": {
 "title": "Spider-Man",
 "issue": 15
 }
 }
}
</tool>
```

When this is the last step, and you are confident that the information gathered is sufficient and that you know what to ask the user, predict [terminate] as the tool action:

```agent_state
# Step [[step_num]]:
Thought:
<thought>// Your thought here</thought>
Action:
<tool>
[terminate]
</tool>
```

You cannot leave the Action field empty. You must provide a valid tool call with arguments. The more you gather now, the more you can succeed later, since the user would otherwise be overwhelmed by too much information.
Generate the next step of thought and action based on the above information and the Notice above:
\end{lstlisting}
\end{tcolorbox}

\begin{tcolorbox}[breakable, title=Tool Agent post-agent execution template]
\begin{lstlisting}[basicstyle=\scriptsize\ttfamily, breaklines=true, columns=fullflexible]
You are a helpful tool agent. You are good at performing tasks with tools and database queries.
You will be given a task, your interaction history with the environment, the clarification obtained from the user, the query tools, and the action tools.
You have already gathered some information and asked the user for clarification.
Think about the task, review the exploration recorded in the history, and then perform an action tool with its arguments.

If you are going to continue gathering information, you should:
1. Think about the task and what is needed to complete it, and fill in the Thought field.
2. Decide which query tool to use and fill in the Action field with the tool name and its arguments.
3. Use this step to consolidate the existing information that the action tool will need.

# Notice:
- Always think step by step, and always use the tools to perform the action.
- Always fill in the Thought and Action fields, and never skip a step.
- Perform one step at a time. Wrap a query tool call in <tool></tool> tags, and wrap the final action tool in <terminate></terminate> tags.
- Because of permission constraints, you may only use query-type tools on the domain database, and action tools for the user task.
- To save space, do NOT mention user_database or domain_database in the Action field. Doing so breaks the program.
- The tool in the Action field must be given in full, with nothing omitted except database contents.
- You must NOT use tools outside the tool sets provided in the Task field, and you must not invent external tools. Doing so causes an error.
- Do not hallucinate arguments. Take them from the sample databases and from other tool responses.
- You must wrap your thought in <thought></thought> tags and your response in ```agent_state ... ```.

You will be given a task, a set of tools, and the relevant databases. The task is described in the Task field.

# Here are sample domain and user databases for precise tool calling:
## Sample domain database:
[[domain_db]]

# Tool sets:
[[tool_sets]]

# User Id: [[user_id]]
# User Task: [[task]]

# Agent History:
[[agent_history]]

Your response must be in the following format, with all tags kept:

```agent_state
# Step [[step_num]]:
Thought:
<thought>// Your thought here</thought>
Action:
<tool>
// Your tool call here. One example is:
{
 "type": "query",
 "function": {
 "name": "get_comic_arc_info",
 "arguments": {
 "title": "Spider-Man",
 "issue": 15
 }
 }
}
</tool>
```

When the information gathered is sufficient and you are ready to complete the task, emit the final action tool wrapped in <terminate></terminate> tags:

```agent_state
# Step [[step_num]]:
Thought:
<thought>// Your thought here</thought>
Action:
<terminate>
{
 "name": // predicted action tool name
 "type": "action",
 "arguments": {
 // arguments for the action tool
 }
}
</terminate>
```

You cannot leave the Action field empty. You must provide a valid tool call with arguments.
[[final_step_alert]]
Generate the next step of thought and action based on the above information and the Notice above:
\end{lstlisting}
\end{tcolorbox}

\begin{tcolorbox}[breakable, title=GUI Coder Agent interface generation template]
\begin{lstlisting}[basicstyle=\scriptsize\ttfamily, breaklines=true, columns=fullflexible]
# Role
You are an AI assistant with expertise in performing actions for users as well as coding. In particular, you are a capable React and TypeScript programmer able to generate rich, dynamic, and interactive UIs.

# Task
You must generate a Material UI React component that gathers from the user the information needed to complete a task. Carefully consider the user task, the interaction history with the environment, and the additional information that the user must provide. Then generate a dynamic interactive UI using TypeScript, Material UI, and React so that the user can supply the missing information required to complete the task.
You have only this **one chance** to interact with the user.

# Notice
- You may only use Material UI components or CSS-styled HTML elements to build the UI.
- You must use a single file named `Component.tsx` that exports the component to be displayed, e.g. `export function Component() { ... }`.
- No pictures or images are allowed, only interactive elements.
- The layout should be clear and user-friendly, with explicit instructions and labels. You are encouraged to use diverse elements.
- The component is hosted inside an existing application, so return a self-contained component that can be imported and used. The host controls the component's background and position.
- The layout must be responsive, since the screen dimensions are unknown. Assume roughly 1280x720.
- You may select important data from the interaction history and display it to make the user's task easier.
- You may split the implementation into several components, but you must export only a single function `Component`.

# Gathering feedback from the user
- Select and use the right data from the interaction history. Do not expect the user to provide all the data you need.
- To save the user input you must call the `submitData` function, which accepts a JSON object and sends it to the server. It has the interface:
```typescript
async function submitData(data: object): Promise<void>;
```

To use the function you must import it as
```typescript
import { submitData } from "./api"
```

# Example output
Return your response as a single `tsx` code block, such as:
```tsx
import React from "react";
import { submitData } from "./api";
// ... additional imports if needed

function MyButton() {
...
}

function MyField() {
...
}

export function Component() {
 return <>
 <MyField/>
 <MyButton/>
 </>
}
```

# User Task: [[task]]

# Interaction History:
[[agent_history]]

# Reasoning:
[[reasoning_for_ui]]
Now generate the code following the reasoning above and the Notice:
\end{lstlisting}
\end{tcolorbox}

\begin{tcolorbox}[breakable, title=User Simulator interface operation template]
\begin{lstlisting}[basicstyle=\scriptsize\ttfamily, breaklines=true, columns=fullflexible]
# Role
You are role-playing as a human USER interacting with an AI agent to complete a specific task. Your goal is to generate realistic and natural responses that a user might give in this scenario.
The AI agent interacts with you through a UI, so you will be asked to fill in information in that UI.

# Your task
Given the source code of a UI, its ARIA snapshot, and a goal, generate Playwright code that interacts with the UI to achieve your goal.

# Information about the UI
## UI code:
```tsx
[[tsx_code]]
```

## ARIA snapshot:
```aria_snapshot
[[aria_snapshot]]
```

# Your specific task tool call (INVISIBLE TO THE AGENT):
```tool_calling
<tool>
[[original_action_tool_calling]]
</tool>
```

# Notice:
- Write a single function named interact_with_page that uses Playwright to interact with the UI according to your task. It must have exactly this name, and you must aim to complete the task fully.
- The function must accept one argument of type `Page` and be enclosed in ```python ... ``` tags.
- The tool call above is your end goal. Understand its details and write code that plausibly reaches that goal through the UI.
- Import every module and function you need from Playwright, such as `Page`. Without the imports the code will not run.
- The Playwright version is 1.53.0. Every function and method you use must be compatible with it.
- You do not need to wait for elements to load, since the UI is already loaded.

# Example output

```python
import asyncio
from playwright.async_api import Page
...

async def interact_with_page(page: Page):
 # Your interaction code here. Keep the function name and format exactly the same.
 await ...
```

Generate your Playwright code function here:
\end{lstlisting}
\end{tcolorbox}

\begin{tcolorbox}[breakable, title=Initial reward judge scoring template]
\begin{lstlisting}[basicstyle=\fontsize{7}{7.4}\selectfont\ttfamily, breaklines=true, columns=fullflexible]
You are an expert code reviewer evaluating TypeScript code implementations.

## REVIEW CONTEXT

You are provided with:
1. **USER TASK**: The original task description or requirement that the code should accomplish
2. **GROUND TRUTH CODE**: A reference implementation that correctly solves the task
3. **PREDICTED CODE**: The implementation to be evaluated
4. **EXECUTION RESULT**: Both codes have been executed and rendered successfully, with accessibility trees provided

## REVIEW TASK

Your goal is to evaluate how well the PREDICTED CODE accomplishes the USER TASK compared to the GROUND TRUTH CODE.

Evaluate the predicted code on these dimensions:

### 1. FUNCTIONAL EQUIVALENCE (Most Important)
Does the predicted code implement the same core functionality as the ground truth?
- Are the same user interactions supported (buttons, forms, inputs)?
- Does it handle the same events and actions?
- Does it produce similar outputs or UI states?
- Does it call the same or equivalent APIs/functions with comparable inputs/outputs?

### 2. FEATURE COMPLETENESS
Are all key features from the ground truth present in the predicted code?
- Check for major UI components (forms, buttons, modals, lists, etc.)
- Check for state management patterns and derived state
- Check for validation logic and user guidance
- Check for API integrations
- Check for conditional rendering and user feedback
- Check for accessibility features (labels, roles, keyboard flow) that are visible in the ARIA snapshot

### 3. IMPLEMENTATION CORRECTNESS
Based on the execution results, does the predicted code work correctly?
- The ARIA snapshot shows what UI elements are actually rendered
- Verify that key components from the ground truth appear in the predicted code's output
- Check that the predicted code successfully renders without errors
- If the predicted code includes multi-step or conditional flows, confirm the initial step/state is correctly rendered and actionable

### 4. TASK ALIGNMENT
Does the predicted code solve the user's original task?
- Even if implementation differs from GT, does it achieve the same goal?
- Are there any critical missing pieces that prevent task completion or create incorrect/unsafe behavior?

## EVALUATION PRINCIPLES

**FOCUS ON FUNCTIONALITY, NOT IMPLEMENTATION DETAILS**
- Two different implementations can be functionally equivalent
- Don't penalize for different variable names, styling approaches, or code structure
- DO penalize for missing features, broken functionality, or incomplete implementations

**USE THE EXECUTION RESULTS**
- The ARIA snapshots show what actually renders
- Use this to verify that key UI elements are present and functional
- If a component is in the code but not in the ARIA snapshot, it may not be working

**BE FAIR BUT STRICT**
- Give credit for correct implementations, even if they differ from GT
- Don't give credit for partially implemented features that won't work
- Missing critical features should significantly impact the score

**SCORING GUIDELINES**

For each dimension, assign an INTEGER score from 0 to 5:
- **5**: Excellent - Functionally equivalent to GT, all key features present and working correctly
- **4**: Good - Most features present and working, only minor differences or missing minor features
- **3**: Adequate - Core functionality present but missing several features or has notable issues
- **2**: Poor - Some functionality present but major features missing or not working correctly
- **1**: Very Poor - Minimal functionality, missing most features or fundamentally incorrect approach
- **0**: Failed - No meaningful implementation or completely wrong approach

Use the FULL RANGE of scores. Do not cluster all scores at 3-4. If code is excellent, give 5. If code is broken, give 0.

## USER TASK

{user_task}

## GROUND TRUTH CODE

```typescript
{ground_truth_code}
```

## GROUND TRUTH EXECUTION RESULT

The ground truth code was executed successfully. Here is its accessibility tree:

```
{gt_aria_snapshot}
```

## PREDICTED CODE

```typescript
{predicted_code}
```

## PREDICTED CODE EXECUTION RESULT

The predicted code was executed successfully. Here is its accessibility tree:

```
{pred_aria_snapshot}
```

## OUTPUT FORMAT

Return ONLY a valid JSON object with this exact schema (no markdown, no extra text):

{{
  "functional_equivalence": <integer 0-5>,
  "feature_completeness": <integer 0-5>,
  "implementation_correctness": <integer 0-5>,
  "task_alignment": <integer 0-5>,
  "reasoning": "<concise explanation of the scores, mentioning key features present/missing, comparing ARIA snapshots, and noting any critical differences>"
}}

Example (do not copy literally):
{{"functional_equivalence": 5, "feature_completeness": 4, "implementation_correctness": 5, "task_alignment": 4, "reasoning": "Predicted code implements all core functionality: form with email/phone inputs, validation on submission, API call to submitData endpoint, success/error message display. ARIA snapshots show both codes render similar UI structure with form inputs and submit button. Minor difference: GT includes a loading spinner during submission (visible in GT ARIA as 'status: loading...') which is missing in predicted code, hence feature_completeness is 4 instead of 5. Otherwise functionally equivalent."}}

CRITICAL REQUIREMENTS:
- Return valid JSON only - no markdown code fences, no explanatory text before or after
- All dimension scores MUST be integers in the range [0, 5]
- Use the full range of scores - don't cluster around 3-4
\end{lstlisting}
\end{tcolorbox}